\let\usbgprimitiveyear\year
\documentclass{ieeeaccess}

\IfFileExists{IEEEtran.cls}{}{}\IfFileExists{spotcolor.sty}{}{}\IfFileExists{figures/logo.png}{}{}\IfFileExists{figures/notaglinelogo.png}{}{}
\IfFileExists{figures/bullet.png}{}{}\IfFileExists{t1-formata.map}{}{}\IfFileExists{t1-giovannistd.map}{}{}\IfFileExists{t1-helvetica.map}{}{}
\IfFileExists{t1-times.map}{}{}\IfFileExists{t1formata.fd}{}{}\IfFileExists{t1giovannistd.fd}{}{}\IfFileExists{t1helvetica.fd}{}{}
\IfFileExists{t1times.fd}{}{}\IfFileExists{t1-formata-bold.tfm}{}{}\IfFileExists{t1-formata-bolditalic.tfm}{}{}\IfFileExists{t1-formata-condmedium.tfm}{}{}
\IfFileExists{t1-formata-condmediumital.tfm}{}{}\IfFileExists{t1-formata-italic.tfm}{}{}\IfFileExists{t1-formata-light.tfm}{}{}\IfFileExists{t1-formata-lightitalic.tfm}{}{}
\IfFileExists{t1-formata-medium.tfm}{}{}\IfFileExists{t1-formata-mediumitalic.tfm}{}{}\IfFileExists{t1-formata-regsymb.tfm}{}{}\IfFileExists{t1-formata-regular.tfm}{}{}
\IfFileExists{t1-giovannistd-bookitalic.tfm}{}{}\IfFileExists{t1-times-bold.tfm}{}{}\IfFileExists{t1-times-bolditalic.tfm}{}{}\IfFileExists{t1-times-italic.tfm}{}{}
\IfFileExists{t1-times.tfm}{}{}\IfFileExists{t1-formata-bold.pfb}{}{}\IfFileExists{t1-formata-bolditalic.pfb}{}{}\IfFileExists{t1-formata-condmedium.pfb}{}{}
\IfFileExists{t1-formata-condmediumital.pfb}{}{}\IfFileExists{t1-formata-italic.pfb}{}{}\IfFileExists{t1-formata-light.pfb}{}{}\IfFileExists{t1-formata-lightitalic.pfb}{}{}
\IfFileExists{t1-formata-medium.pfb}{}{}\IfFileExists{t1-formata-mediumitalic.pfb}{}{}\IfFileExists{t1-formata-regsymb.pfb}{}{}\IfFileExists{t1-formata-regular.pfb}{}{}
\IfFileExists{t1-giovannistd-bookitalic.pfb}{}{}\IfFileExists{t1-times-bold.pfb}{}{}\IfFileExists{t1-times-bolditalic.pfb}{}{}\IfFileExists{t1-times-italic.pfb}{}{}
\IfFileExists{t1-times.pfb}{}{}
\usepackage[T1]{fontenc}
\usepackage{amsmath,amssymb}
\usepackage{graphicx}
\usepackage{booktabs}
\usepackage{tabularx}
\usepackage{array}
\usepackage{enumitem}
\usepackage{xcolor}
\usepackage{colortbl}
\usepackage{float}
\let\usbgclassyear\year\let\year\usbgprimitiveyear
\usepackage{tikz}
\usetikzlibrary{arrows.meta,positioning,shapes.geometric,calc,fit}
\let\year\usbgclassyear
\makeatletter\edef\usbgspot{\the\PANTONE}\expandafter\pgfutil@addpdfresource@colorspaces\expandafter{\usbgspot}\makeatother
\usepackage{url}
\floatstyle{boxed}\newfloat{usbgbox}{tbp}{lob}\floatname{usbgbox}{Box}
\makeatletter
\newcommand{\floatc@usbglisting}[2]{{\color{accessblue}\figcapheadfont\MakeUppercase{#1}.\ \ }{\figcapfont #2}\par}
\newcommand{\fs@usbglisting}{\def\@fs@cfont{}\let\@fs@capt\floatc@usbglisting
  \def\@fs@pre{}\def\@fs@post{}\def\@fs@mid{\vspace{5pt}}\let\@fs@iftopcapt\iftrue}
\makeatother
\floatstyle{usbglisting}\newfloat{listing}{tbp}{lol}\floatname{listing}{Listing}
\usepackage{listings}
\lstdefinestyle{usbgpy}{language=Python, basicstyle=\ttfamily\fontsize{7.5}{9.4}\selectfont, columns=fixed, keepspaces=true,
  showstringspaces=false, numbers=left, numberstyle=\fontsize{6}{7}\selectfont\color{black!55}, numbersep=5pt,
  frame=single, framerule=0.4pt, rulecolor=\color{black!35}, backgroundcolor=\color{black!4},
  keywordstyle=\bfseries, commentstyle=\color{black!60}, stringstyle=\color{black},
  xleftmargin=1.7em, framexleftmargin=1.5em, aboveskip=0pt, belowskip=0pt, escapeinside={(*@}{@*)}}
\floatstyle{plain}
\usepackage{xr-hyper}
\usepackage[hidelinks]{hyperref}

\providecommand{\theHlisting}{\arabic{listing}}
\graphicspath{{figures/}}
\makeatletter\providecommand{\ContinuedFloat}{\addtocounter{\@captype}{-1}}\makeatother
\IfFileExists{supp_labels.tex}{\input{supp_labels}}{\typeout{main.tex: supp_labels.tex missing; run build.sh}}

\begin{document}
\history{Date of publication xxxx 00, 0000, date of current version xxxx 00, 0000.}
\doi{10.1109/ACCESS.2024.0429000}

\title{USB Gym: A Gymnasium Testbed for Closed-Loop Ultrasound View Matching}

\author{\uppercase{Thanapong Intharah}\authorrefmark{1}, \IEEEmembership{Member, IEEE},
\uppercase{Zhichao Gao}\authorrefmark{2},
\uppercase{Yiliang Liu}\authorrefmark{2}, AND
\uppercase{Hao Dong}\authorrefmark{2}}

\address[1]{Visual Intelligence Laboratory, Department of Statistics, Faculty of Science, Khon Kaen University, Khon Kaen 40002, Thailand}
\address[2]{School of Computer Science, Peking University, Beijing 100871, China}

\tfootnote{This work was supported in part by the NSRF via the Program Management Unit for Human Resources \& Institutional Development, Research and Innovation under Grant B13F680141, and in part by the NSRF via the Research and Innovation Acceleration Agency for Competitiveness and Area Development (RCAD) (Program Management Unit for Frontier Brainpower and Future Industries) under Grant B13T690002-4.}

\markboth
{T. Intharah \headeretal: USB Gym: A Gymnasium Testbed for Closed-Loop Ultrasound View Matching}
{T. Intharah \headeretal: USB Gym: A Gymnasium Testbed for Closed-Loop Ultrasound View Matching}

\corresp{Corresponding author: Thanapong Intharah (e-mail: thanin@kku.ac.th).}

\begin{abstract}
Our main goal is ultrasound guidance: moving a probe until its image matches a target view.
Published methods for each part of a guide are evaluated in separate simulators, so we asked whether these methods, combined into one guide, are good enough.
We built USB Gym, a testbed with six replaceable parts in two groups.
Its standard engine (anatomy, renderer and a network drawing organ textures from tissue labels) simulates ultrasound from public patient computed tomography and was held fixed.
Its guide (organ segmenter, controller and reward) received 20 published methods, each run on the same 60 episodes of six held-out patients.
In each guide part, a published method was used instead of ours only if detectably better on these patients.
The better of this combination and our original guide on 20 new scans is the benchmark baseline.
A pose oracle, reading the exact probe pose available only in simulation, set the reference.
The results are exploratory.
On these scans, the benchmark baseline ended 13.7~mm from the target view, from an average start of 16.0~mm, 16\% of the progress of the pose oracle, at 1.5~mm.
It is our guide with an image-similarity reward, structural similarity, the only published method that was slightly but detectably better among the results known when the combination was fixed.
Combining untuned published methods with ours did not guide the probe near the target view, and the task remains under-explored.
The testbed accepts new methods; no guide was tested on hardware.
\end{abstract}


\titlepgskip=-21pt

\maketitle

\section{Introduction}\label{sec:intro}

Our main goal is ultrasound guidance for view matching: a closed loop that moves a probe until its image matches a target view.
Such a guide combines several parts, among them a probe-navigation policy, an ultrasound segmenter and, in simulation, a B-mode renderer, but the groups that propose them rarely share a closed loop in which to compare them~\cite{vonhaxthausen2021}.
Without a shared loop, their errors on view matching share no scale.

To the best of our knowledge, no public testbed lets published methods replace any of several loop components and be evaluated against a privileged oracle ladder on identical episodes.
We therefore asked whether existing published components, combined into one guide, are good enough for this task, and built a testbed into which research components plug and are scored.
Here, a guide is good enough, and solves the task, if it ends as close to the target as the privileged pose oracle without privileged input at run time; we report how far each pipeline remains from it.

USB Gym (UltraSound B-mode Gym) is that testbed, built on Gymnasium~\cite{gymnasium} and patient computed tomography (CT)~\cite{totalseg}.
Its standard simulation engine renders patient CT with a physics renderer and the SonoSPADE texture stage, and its guide has three plug slots: mask source, controller and reward (Fig.~\ref{fig:pipeline}, Table~\ref{tab:slots}). We plugged 20 published methods into the guide's slots, each in place of its slot's reference component, on that component's 60 held-out episodes; seven more, in the engine's three slots, are in \ref{app:decisions}.
An oracle ladder runs from a pose oracle, which reads privileged simulator state that no physical rig provides, down to a static probe (Section~\ref{sec:ladder}).
SonoSPADE~\cite{sonospade} is the engine's texture stage, used unchanged and retrained once.

Here, existing published components, untuned, do not close the gap to the pose oracle, whether plugged one slot at a time or combined with our published pipeline~\cite{usbgym} (Section~\ref{sec:plug-best}).
From a 16.0~mm start on six held-out patients, the pose oracle ends at 1.5~mm, but the CT-mask servo, steering by the CT's true organ masks (privileged CT masks), ends at 11.6 to 11.8~mm.
No published controller in place of ours, given the same input, ends detectably closer (exploratory).
Steering by a segmenter's masks, the learned-mask servo ends above the start with the ladder's three other texture settings.
With SonoSPADE, the standard stage and the only one reading the CT label, it ends at 13.1~mm, 3.4 to 3.8~mm closer than with any of them.

Only the pre-specified analyses of the reference pipeline and of real frames are primary;
every other result, including the comparison of published methods, is exploratory (Section~\ref{sec:stats}).
All closed-loop results come from simulation with rigid tissue, never from a probe, phantom or robot, and we make no claim about clinical usability.

The contributions are:
\begin{enumerate}[leftmargin=*]
\item \textbf{A closed-loop testbed on patient CT} (Sections~\ref{sec:testbed} to~\ref{sec:results-ladder}), with privileged state kept out of the policy's observation.
\item \textbf{A slot-by-slot reference set for the guide} (Section~\ref{sec:pluggability}): every method's error on the standard engine, paired with its slot's reference component.
\item \textbf{Controls and replication} (Section~\ref{sec:checks}): a positive control, label-input controls, an offline real-frame score and two replication cohorts in simulation.
\item \textbf{A benchmark baseline} (Box~\ref{box:reading}; Section~\ref{sec:plug-best}): to test the testbed, slot components chosen on six held-out patients are combined and tested on 20 new CTs (exploratory).
On the standard engine, it is that combination, 13.70~mm from the target (pose oracle 1.47~mm; conditional on this renderer).
\end{enumerate}

The rest of the paper is organized as follows.
Section~\ref{sec:related} reviews ultrasound simulation, learned probe navigation and runnable testbeds.
Section~\ref{sec:testbed} describes the testbed, and Section~\ref{sec:using} its interface, plug protocol and compute cost.
Section~\ref{sec:reference} gives the reference pipeline and the evaluation design.
Sections~\ref{sec:results-ladder} and~\ref{sec:pluggability} report the reference pipeline and the published methods in the guide's slots, and Section~\ref{sec:checks} the controls, sensitivity analyses, replication cohorts and the offline evaluation on real frames.
Section~\ref{sec:discussion} discusses the findings and limitations, and Section~\ref{sec:conclusion} concludes.
The supplementary material holds Supplementary Appendices A to C (evaluation details, every decision of the plug study and the leaderboard of plugged methods) and Supplementary Sections S1 to S7 (renderer, further analyses, implementation and compute details).

\section{Related work}\label{sec:related}

\subsection{Ultrasound simulation}
Ultrasound simulators turn labeled anatomy into B-mode frames for a probe pose, from CT on a graphics processing unit (GPU)~\cite{kutter2009}, by convolutional ray tracing of a labeled CT volume~\cite{salehi2015} or by simulating the acoustic field, as Field II~\cite{jensen1992field}, k-Wave~\cite{treeby2010kwave} and SIMUS~\cite{garcia2022simus,cigier2022simus} do; they leave the closed loop to the user.
USB Gym's renderer slot runs SIMUS and the original code of LOTUS, which learns such a ray model jointly with a segmentation task~\cite{velikova2023lotus} (\ref{sec:plug-renderer}).

\subsection{Learned texture}
Learned texture stages move simulated frames toward real-scan appearance, and USB Gym's texture slot accepts one.
CACTUSS translates real B-mode images into an intermediate representation rendered from CT labels~\cite{cactuss}.
Vitale et al.\ added CycleGAN texture to ray-traced abdominal simulations~\cite{cyclegan,vitale}, and SonoSPADE conditions texture on a tissue label map~\cite{spade,sonospade}.

\subsection{Learning probe navigation}
Learned probe-navigation methods, reviewed in two surveys~\cite{vonhaxthausen2021,jiang2023}, are each trained and evaluated in their own environment.
Some learn from recorded scans: Droste et al.\ predicted probe motions offline from 464 tracked clinical scans~\cite{droste2020}, and Li et al.\ steered a virtual probe through 41 reconstructed spine volumes, rewarded by its distance to a goal pose~\cite{li2021}.
Others train on simulated images: Amadou et al.\ trained goal-conditioned agents on images path-traced offline from 929 patient CTs~\cite{amadou2024}.
Hu et al.\ trained a deep Q-network on CT reslices synthesized into ultrasound by Field II, CycleGAN or U-GAT-IT~\cite{hu2024spie}, and their later agent, rewarded by CT labels, reached 75.8\% success on a phantom and 74.0\% on a cadaver~\cite{hu2026tmrb}.
Han et al.\ imitated a virtual expert that reads the true pose, in a ray-casting simulator with a CUT texture stage on magnetic resonance imaging (MRI) label maps~\cite{han2026}.
Qian et al.\ compared nine image generators behind one soft actor-critic (SAC) policy~\cite{qian2026}; one of their policies, trained only in simulation, succeeded in 390 of 400 zero-shot deployments on a robot over a phantom.
As of October 2026, we found no runnable environment released with these seven method papers.

\subsection{Visual servoing}
Ultrasound visual servoing maps image features to probe velocity: Mebarki et al.\ controlled six degrees of freedom from organ image moments~\cite{mebarki2010}, and Krupa et al.\ used speckle to stabilize a B-mode view against tissue motion~\cite{krupa2009}.
USB Gym's reference controller starts from such a map: its servo base, an image servo, fits the map by least squares rather than deriving it (Box~\ref{box:reading}).

\subsection{Runnable testbeds}
Simulation testbeds let robot-learning groups evaluate methods in one fixed environment.
Table~\ref{tab:capability} lists the ultrasound environments that a researcher can run and plug a method into.
Hase et al.\ released a Gym environment and its data, frames recorded from 34 volunteers~\cite{hase2020}.
VesNet-RL released an environment of simulated binary vessel images~\cite{bi2022vesnet}.
Its agent segments real images into such masks before acting.
SonoGym brings such testbeds to robotic ultrasound on NVIDIA Isaac Lab, with orthopedic navigation, bone-surface reconstruction and spinal surgery tasks~\cite{sonogym,isaaclab}, and renders CT anatomy by ray tracing or with the paired translator pix2pix~\cite{sonogym,pix2pix}.
NVIDIA's Isaac for Healthcare adds ultrasound robot workflows and a ray-traced ultrasound sensor on Isaac Sim~\cite{i4h}.
USB Gym builds on Gym, Gymnasium and MuJoCo~\cite{openaigym,gymnasium,mujoco}, and its privileged channel lets an imitation teacher see state that the policy cannot, as in Learning by Cheating~\cite{chen2020lbc}.

\begin{table*}[tp]
\caption{Runnable ultrasound navigation environments by publication year, as each source states them in October 2026.
Privileged state: simulator state that a physical rig lacks, used by a reward, a teacher or the evaluation.
$^{\dagger}$Privileged structure: the contact plane through the target window face and the start lift onto it (Section~\ref{sec:contact}).
Components swapped: parts swapped between runs, with the number of published methods run in them.}
\label{tab:capability}
\centering
\renewcommand{\baselinestretch}{1}\scriptsize
\setlength{\tabcolsep}{2pt}
\begin{tabularx}{\linewidth}{@{}>{\raggedright\arraybackslash}p{0.10\linewidth}*{5}{>{\raggedright\arraybackslash}X}@{}}\toprule
Environment & Image source & Closed loop & Components swapped (published methods) & Privileged state used & Robot or phantom test \\
\midrule
Hase et al., 2020~\cite{hase2020} & Recorded frames of 34 volunteers on a grid & Yes, over the grid & Policy: the authors' DQN variants & Grid position, in the reward & None reported \\
VesNet-RL, 2022~\cite{bi2022vesnet} & Simulated binary vessel images & Yes, in simulation & None described & Centerline distance, in the reward & Robot arm on vascular phantoms \\
SonoGym, 2025~\cite{sonogym} & Ray tracing or pix2pix on TotalSegmentator CT & Yes, in simulation & Policy (5); renderer (2) & Goal pose in the reward; true state for the expert & None reported \\
Isaac for Healthcare, 2025~\cite{i4h} & OptiX ray tracing of labeled surface meshes & Yes; camera images or poses as input & Policy (2) & Target pose in a task's observation and reward & None described \\
USB Gym (this work) & Physics render of CT label slices; optional texture stage & Yes, in simulation & Six slots (27; Table~\ref{tab:slots}) & Pose and CT masks, as rungs and in training; target-anchored contact$^{\dagger}$ & None \\
\bottomrule
\end{tabularx}
\end{table*}

Only USB Gym runs published methods in six components, the mask source and reward among them, and evaluates each on identical episodes against the privileged references of an oracle ladder (Section~\ref{sec:ladder}).
Elsewhere in Table~\ref{tab:capability}, published methods occupy at most the policy and the image source.

\section{USB Gym: testbed design}\label{sec:testbed}
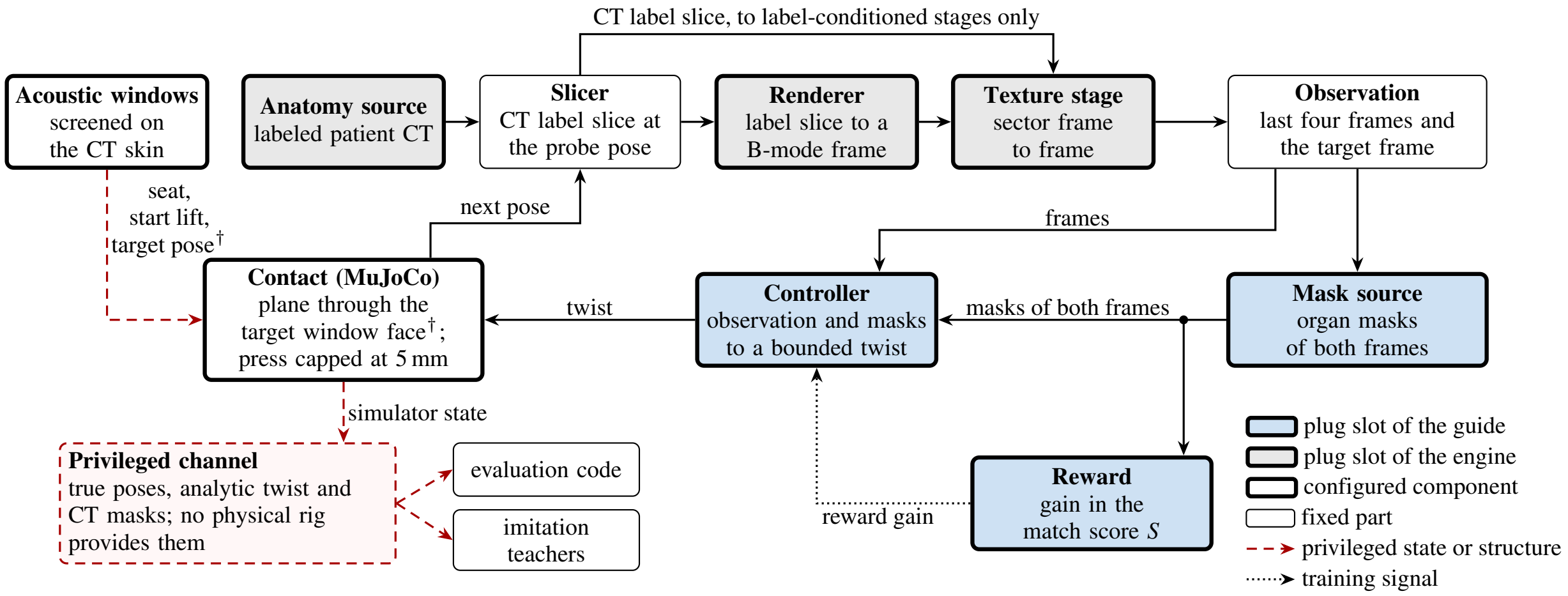
\begin{figure*}[t]
\centering
\begin{tikzpicture}[
  font=\linespread{1}\scriptsize,
  every node/.append style={execute at begin node={\hyphenpenalty=10000\exhyphenpenalty=10000}},
  >={Stealth[length=1.5mm,width=1.2mm]},
  part/.style={draw, rounded corners=2pt, align=center, inner sep=2.5pt, minimum height=0.95cm, fill=white,
               text width=#1},
  part/.default=1.9cm,
  slot/.style={part=#1, very thick, fill=black!9},
  slot/.default=1.9cm,
  gslot/.style={part=#1, very thick, fill={rgb,255:red,206;green,225;blue,243}},
  gslot/.default=1.9cm,
  conf/.style={part=#1, very thick},
  conf/.default=1.9cm,
  flow/.style={->, semithick},
  priv/.style={->, semithick, densely dashed, draw=red!65!black},
  train/.style={->, semithick, densely dotted},
  elab/.style={font=\linespread{1}\scriptsize, inner sep=1pt, align=center},
]
\node[conf] (win) at (1.1,0) {\textbf{Acoustic windows}\\ screened on the CT skin};
\node[slot] (ana) at (3.55,0) {\textbf{Anatomy source}\\ labeled patient CT};
\node[part] (sli) at (6.0,0) {\textbf{Slicer}\\ CT label slice at the probe pose};
\node[slot] (ren) at (8.45,0) {\textbf{Renderer}\\ label slice to a B-mode frame};
\node[slot] (tex) at (10.9,0) {\textbf{Texture stage}\\ sector frame to frame};
\node[part=2.5cm] (obs) at (14.05,0) {\textbf{Observation}\\ last four frames and the target frame};
\node[conf=2.7cm] (con) at (3.55,-2.05) {\textbf{Contact (MuJoCo)}\\ plane through the target window face$^{\dagger}$; press capped at 5\,mm};
\node[gslot=2.3cm] (ctl) at (8.45,-2.05) {\textbf{Controller}\\ observation and masks to a bounded twist};
\node[gslot=2.5cm] (msk) at (14.05,-2.05) {\textbf{Mask source}\\ organ masks of both frames};
\node[gslot=2.3cm] (rew) at (11.3,-3.95) {\textbf{Reward}\\ gain in the match score $S$};
\node[draw=red!65!black, densely dashed, semithick, rounded corners=2pt, fill=red!3, align=left, text width=3.3cm,
      inner sep=2.5pt] (priv) at (2.35,-3.95)
  {\textbf{Privileged channel}\\ true poses, analytic twist and CT masks; no physical rig provides them};
\node[part=1.75cm, minimum height=0.55cm] (eva) at (5.65,-3.6) {evaluation code};
\node[part=1.75cm, minimum height=0.55cm] (tea) at (5.65,-4.33) {imitation teachers};
\node[align=left, anchor=west, inner sep=1pt] (legend) at (12.85,-3.95) {\tikz[baseline=-0.6ex]\draw[very thick, rounded corners=1pt, fill={rgb,255:red,206;green,225;blue,243}] (0,-0.09) rectangle (0.5,0.09); plug slot of the guide\\[1pt]
  \tikz[baseline=-0.6ex]\draw[very thick, rounded corners=1pt, fill=black!9] (0,-0.09) rectangle (0.5,0.09); plug slot of the engine\\[1pt]
  \tikz[baseline=-0.6ex]\draw[very thick, rounded corners=1pt] (0,-0.09) rectangle (0.5,0.09); configured component\\[1pt]
  \tikz[baseline=-0.6ex]\draw[rounded corners=1pt] (0,-0.09) rectangle (0.5,0.09); fixed part\\[1pt]
  \tikz[baseline=-0.6ex]\draw[priv] (0,0) -- (0.5,0); privileged state or structure\\[1pt]
  \tikz[baseline=-0.6ex]\draw[train] (0,0) -- (0.5,0); training signal};
\draw[flow] (ana) -- (sli);
\draw[flow] (sli) -- (ren);
\draw[flow] (ren) -- (tex);
\draw[flow] (tex) -- (obs);
\draw[flow] (sli.north) -- ++(0,0.42) -| node[elab, above, pos=0.25] {CT label slice, to label-conditioned stages only} (tex.north);
\draw[flow] (obs) -- (msk);
\draw[flow] (13.2,0 |- obs.south) -- (13.2,-1.12) -- node[elab, above, pos=0.5] {frames} (9.1,-1.12) -- (9.1,0 |- ctl.north);
\draw[flow] (msk.west) -- node[elab, above, pos=0.55] {masks of both frames} (ctl.east);
\fill (12.25,-2.05) circle (1.3pt);
\draw[flow] (12.25,-2.05) -- (12.25,0 |- rew.north);
\draw[train] (rew.west) -| node[elab, below, pos=0.3] {reward gain} (ctl.south);
\draw[flow] (ctl.west) -- node[elab, above] {twist} (con.east);
\draw[flow] (4.45,0 |- con.north) |- node[elab, above, pos=0.75] {next pose} (6.0,-1.05) -- (sli.south);
\draw[priv] (win.south) |- node[elab, right, pos=0.17] {seat,\\ start lift,\\ target pose$^{\dagger}$} (con.west);
\draw[priv] (con.south) -- node[elab, right] {simulator state} (con.south |- priv.north);
\draw[priv] (priv.east) -- (eva.west);
\draw[priv] (priv.east) -- (tea.west);
\end{tikzpicture}
\caption{The USB Gym closed loop. Each step slices the patient's labeled CT at the probe pose, renders a frame and applies a texture stage. A mask source labels the organs of the observation and the target frame. The controller's twist moves the probe, a lagged position servo, against a contact plane through the target window face. This target-anchored contact and the start lift onto it are privileged structure ($^{\dagger}$, red dashed arrow). Shaded boxes are the six plug slots: the three of the guide (blue), which this article compares, and the three of the standard engine (gray), which it holds fixed. White boxes with a heavy outline are configured components (Table~\ref{tab:slots}). The dashed privileged channel feeds evaluation code and imitation teachers only.}\label{fig:pipeline}\label{fig:loop}\end{figure*}

\subsection{Design goals, loop and task}\label{sec:gym}\label{sec:task}
USB Gym keeps every component of a closed ultrasound loop fixed except the one a user replaces, so a change in final error can be traced to that component.
In each step (Fig.~\ref{fig:loop}), the pose selects a CT label slice, which the simulation engine, a renderer and a texture stage, turns into the observation.
A mask source labels the organs of the observation and target for the reward to score.
The controller outputs a twist, a velocity of three translations and three rotations, which MuJoCo~\cite{mujoco} executes against contact.

The task is a Gymnasium~\cite{gymnasium} environment in which the probe moves from a perturbed start to match a target view.
An acoustic window is a skin pose from which the beam reaches the target organ largely free of bone, lung and bowel gas.
Each target is the view at a screened liver or kidney window.
Its start is drawn uniformly within $\pm 20$~mm lateral and elevational, $\pm 10$~mm axial, $\pm 18^\circ$ of tilt and $\pm 4^\circ$ of roll.
Each step commands at most 3~mm along and 0.05~rad about each axis.
The observation contains the last four $128 \times 128$ frames, the target frame and optional organ masks.
Episodes have no terminal state and are truncated at 12 steps.
The task is therefore local re-acquisition near a known window: finding a window from afar and deciding when to stop are out of scope.

\subsection{Slots and interfaces}\label{sec:slots}
Table~\ref{tab:slots} lists the eight configurable components, the six plug slots among them, and what a replacement of each provides (slot terms and rules for reading the results: Box~\ref{box:reading}). The anatomy source, renderer and texture stage form the standard simulation engine, held at their reference components while the guide's slots, mask source, controller and reward, are compared (Section~\ref{sec:pluggability}); the engine's slots stay pluggable (Supplementary Table~\ref{tab:pluggability-engine}).

\makeatletter
\@ifundefined{c@usbgbox}{\floatstyle{boxed}\newfloat{usbgbox}{tbp}{lob}\floatname{usbgbox}{Box}\floatstyle{plain}}{}
\makeatother
\begin{usbgbox*}[tp]
\centering
\renewcommand{\baselinestretch}{1}\footnotesize
\textbf{(a) Terms.}\par\smallskip
\begin{tabularx}{\linewidth}{@{}>{\raggedright\arraybackslash}p{3.35cm} >{\raggedright\arraybackslash}X@{}}
\toprule
Term & Definition \\
\midrule
Configurable component & A loop part that a user can replace; Table~\ref{tab:slots} lists eight. \\
Slot & A configurable component into which published methods were plugged; six of the eight. \\
Standard engine; guide & The engine is the physics renderer, the SonoSPADE texture stage and the reference CT labels; the guide is the mask source, controller and reward. All six slots are pluggable; Section~\ref{sec:pluggability} compares the guide's. \\
Reference component & A component of the primary results, ours or a prior one, against which a slot's plugged methods are compared. \\
Family & Pre-specified comparisons within which Holm's adjustment runs; in a slot's family, each tests a plugged method. \\
Secondary comparison & A reported comparison outside every family, with a $t$ confidence interval (CI) and no decision. \\
Privileged component ($^{\dagger}$) & A component that reads simulator state that no physical rig provides. \\
Target-anchored contact ($^{\dagger}$) & Contact plane through the target window face, and the start lift onto it; privileged structure. \\
Privileged channel & The separate simulator output holding the true poses, the analytic twist and the CT masks. \\
Analytic twist & The clipped corrective twist from the true poses; the pose oracle executes it. \\
Oracle ladder; rung & References that see less and less simulator state on identical episodes; each is a rung. \\
Match score $S$; the paper's $S$ & Eq.~\eqref{eq:score}: weighted organ-mask overlap plus in-sector appearance; it uses no pose. \\
Servo base & Our image servo: a least-squares linear map from mask moments to a twist. \\
Reference pipeline; authors' original pipeline & Our published pipeline~\cite{usbgym}: one of the four reference texture stages, the stage-matched U-Net mask and the reference controller, trained with $S$; every other slot at its reference component (Table~\ref{tab:slots}). \\
Label-conditioned stage; label-conditioned frames & A texture stage that also reads the CT label slice, as SonoSPADE does; the frames are its output. Every frame, physics included, is first rendered from that slice. \\
SonoSPADE; per-tissue stage & The published label-conditioned texture stage of~\cite{sonospade}, used unchanged; the standard engine's texture stage. \\
Oracle-to-CT-mask gap & Final error of the CT-mask servo minus that of the pose oracle. \\
CT-to-learned-mask gap & Final error of a learned-mask servo minus that of the CT-mask servo. \\
Benchmark baseline & On the standard engine (physics frames: \ref{app:decisions-combination}), the better of two pipelines on the 20 cohort CTs: the authors' original one or, where it tests lower, the combination keeping the authors' component in each slot unless a decision on the six held-out patients found a published one lower (Section~\ref{sec:plug-best}). Others are not ranked against it. \\
Look-ahead ascent & A privileged rung that simulates candidate twists and executes the one with highest CT-mask $S$. \\
Planted deficit & A known lateral bias added to executed twists, to test whether REINFORCE removes it. \\
MDD & Minimum detectable difference: smallest true paired difference a two-sided paired $t$-test detects at power 0.8. \\
TOST & Two one-sided tests of equivalence within a preset margin, here $\pm 1$~mm. \\
\bottomrule
\end{tabularx}
\par\medskip
\textbf{(b) How to read the results.}
\begin{enumerate}[leftmargin=*,itemsep=1pt,topsep=2pt,label=\arabic*.]
\item The primary results are the pre-specified analyses of the reference pipeline with its four reference texture stages (Section~\ref{sec:results-ladder} and \ref{app:eval}) and the primary real-frame results of Section~\ref{sec:results-real} (every stage: Supplementary Table~\ref{tab:real-cells-all}). Every other result, the comparison of published methods, the controls and the replication cohorts included, is exploratory; a pre-specified rule gives it a decision word, and an estimation-only result has none (Supplementary S7). New analyses of data already seen are labeled post hoc, and descriptive results are counts or summaries with no test. Sections~\ref{sec:pluggability} and~\ref{sec:checks-repeat} to~\ref{sec:checks-cohort} and \ref{app:decisions} report only exploratory results, so there a result is labeled only when it is also post hoc, estimation only, secondary or descriptive; values of the six held-out patients quoted from Section~\ref{sec:results-ladder} keep their primary status.
\item The patient is the unit of inference, with $n = 6$ (the six held-out patients, also the original six; TRAIN: the 13 training patients). On the 20 replication CTs (also new, further or cohort CTs) and the 30 AMOS CTs, the CT is the unit. Seeds index trained artifacts, and every contrast is conditional on them.
\item Brackets give a 95\% CI of the per-patient or per-CT paired difference: a patient-cluster bootstrap CI for primary results and a paired $t$ CI for exploratory results. $k$/$n$ counts the patients or CTs that share the sign of the mean.
\item A primary contrast is a difference when its CI excludes 0 and at least 5 of 6 patients share its sign. An exploratory contrast is a difference when its Holm-adjusted paired $t$-test within its family gives $p < 0.05$ and at least 5 of 6 patients share its sign. On the 20 replication and 30 AMOS CTs, a sign-flip test with Holm decides, without a sign count. Every ``lower'' decision against one of our reference components is reported in the main text.
\item A result with no detected difference does not show equivalence. The post hoc MDD gives each slot's resolution; only a TOST shows equivalence, named if post hoc.
\item Privileged rows carry a dagger in every table, and secondary and estimation-only rows never carry a decision. Except in the skin-tangent contact experiment, every absolute error is conditional on the target-anchored plane; contact-off rows keep the start lift and the 5~mm press clamp below that plane. Where no other reason is given, a run that is not reported did not pass the provenance checks.
\item \ref{app:decisions} gives every pre-specified decision in full, keyed to the rows of Table~\ref{tab:pluggability} and Supplementary Table~\ref{tab:pluggability-engine}.
\end{enumerate}
\caption{Terms and reading rules.}
\label{box:reading}
\end{usbgbox*}

\begin{table*}[tp]
\caption{Configurable components of USB Gym: six plug slots, and the acoustic windows and contact, configured but not plugged. The anatomy source, renderer and texture stage form the standard simulation engine (results: Supplementary Table~\ref{tab:pluggability-engine}); the mask source, controller and reward form the guide (Table~\ref{tab:pluggability}). Interface: what a replacement provides. A slot's pre-specified comparisons are made against its first reference component. Plugged: the published methods placed in the slot, led by their count, which takes each method once whatever its backbone, box prompt, pretrained weights or residual form, and leaves out ablations of $S$; the six counts sum to 27. Box~\ref{box:reading} defines the entry types. $^{\dagger}$Privileged: a component that reads simulator state, or the target-anchored contact. $^{\star}$Specified after the first families were fixed, in families of its own. $^{\mathrm{R}}$The authors' reimplementation of the published method. NCC: normalized cross-correlation. SSIM, NMI, LPIPS: structural similarity, normalized mutual information, learned perceptual image patch similarity. BC: behavior cloning. CEM: cross-entropy method. PPO: proximal policy optimization. TD3: twin delayed deep deterministic policy gradient. A2C: advantage actor-critic. DQN: deep Q-network.}
\label{tab:slots}\label{tab:extensions}\centering
\renewcommand{\baselinestretch}{1}\scriptsize
\begin{tabularx}{\textwidth}{@{}>{\raggedright\arraybackslash}p{1.45cm} >{\raggedright\arraybackslash}X >{\raggedright\arraybackslash}p{2.75cm} >{\raggedright\arraybackslash}p{5.3cm} >{\raggedright\arraybackslash}p{2.6cm}@{}}
\toprule
Component & Interface & Reference components & Plugged & Not evaluated, and why \\
\midrule
Anatomy source & a labeled CT volume per patient & TotalSegmentator dataset masks~\cite{totalseg}: 13 training and 6 held-out cases & 1 method. Family$^{\star}$: TotalSegmentator v2 labels of the same CTs, fast mode & none \\ \addlinespace[3pt]
Acoustic windows & screening rules; target organs & screened windows; liver and kidney targets & none & none \\ \addlinespace[3pt]
Renderer & a label slice to a frame in $[0,1]$; physics-renderer parameters & physics renderer (authors'); DR off except in one ablation & 2 methods. Family: LOTUS forward model, original code, published default parameters~\cite{velikova2023lotus}; the authors' reimplementation of it. Secondary: SIMUS~\cite{garcia2022simus,cigier2022simus}, seed 0 & Field II~\cite{jensen1992field}: needs MATLAB. k-Wave~\cite{treeby2010kwave}: needs an NVIDIA GPU, or hours per frame without one \\ \addlinespace[3pt]
Texture stage & a frame to a frame; a label-conditioned stage also reads the label slice & SonoSPADE~\cite{sonospade} (authors', method unchanged). Secondary: physics, CUT~\cite{cut}, CycleGAN~\cite{cyclegan} & 4 methods. Family: FastCUT~\cite{cut}$^{\mathrm{R}}$, DCLGAN~\cite{han2021dclgan}$^{\mathrm{R}}$, UNSB~\cite{kim2024unsb}$^{\mathrm{R}}$, global histogram matching~\cite{gonzalez2018}$^{\mathrm{R}}$ & Echo from Noise~\cite{stojanovski2023echo}, SDM~\cite{zhou2022sdm}: no public abdominal weights. ConPres~\cite{tomar2021conpres}: no public code \\ \addlinespace[3pt]
Mask source & a frame to an organ mask & U-Net~\cite{ronneberger2015unet} of the same stage and seed (authors'). Secondary: CT label$^{\dagger}$ & 7 methods. Family: LR-ASPP~\cite{howard2019mobilenetv3}, DeepLabV3~\cite{chen2017deeplabv3} on MobileNetV3-L or ResNet-50~\cite{he2016resnet} and FCN~\cite{long2015fcn}, from scratch; nnU-Net~\cite{isensee2021nnunet} 2D, 2 epochs; SAM~\cite{kirillov2023sam} and MedSAM~\cite{ma2024medsam} with U-Net boxes. Family$^{\star}$: SAMUS~\cite{lin2024samus} with U-Net boxes; the torchvision networks with ImageNet-pretrained backbones. Secondary: SAM, MedSAM and SAMUS with CT-label boxes$^{\dagger}$ & UltraSam~\cite{meyer2026ultrasam}, USFM~\cite{jiao2024usfm}: need an NVIDIA GPU \\ \addlinespace[3pt]
Contact & probe contact in MuJoCo, on or off & on; plane through the target window face, with the start lift onto it$^{\dagger}$; press capped at 5~mm & none & none \\ \addlinespace[3pt]
Controller & any map from the observation to a normalized twist & REINFORCE residual over the servo base, imitation start (authors'). Secondary: servo base alone; pose oracle$^{\dagger}$ & 8 methods. Family: PPO~\cite{schulman2017ppo}, SAC~\cite{haarnoja2018sac}, TD3~\cite{fujimoto2018td3} and A2C~\cite{mnih2016a3c} residuals, PPO and DQN~\cite{mnih2015dqn} with 13 moves from scratch, in Stable-Baselines3~\cite{sb3}; BC~\cite{pomerleau1991}$^{\mathrm{R}}$ and DAgger~\cite{ross2011dagger}$^{\mathrm{R}}$ from the pose-oracle teacher, residual and from scratch. Secondary: CEM look-ahead~\cite{rubinstein1999cem}$^{\dagger,\mathrm{R}}$ & SonoGym agents~\cite{sonogym}: need an NVIDIA RTX GPU \\ \addlinespace[3pt]
Reward & a frame pair to a score; the weights of $S$ & the match score $S$ (authors'): weights 0.6 and 0.4, action cost 0.003 & 5 methods. Family: SSIM~\cite{wang2004ssim}, NMI~\cite{studholme1999nmi}$^{\mathrm{R}}$. Family$^{\star}$: LPIPS~\cite{zhang2018lpips}. Secondary: organ overlap only, NCC only, whole-sector NCC; pose distance$^{\dagger,\mathrm{R}}$~\cite{sonogym}; pose distance plus organ area$^{\dagger,\mathrm{R}}$~\cite{bi2022vesnet} & none \\
\bottomrule
\end{tabularx}
\end{table*}

\subsection{Anatomy, windows and contact}\label{sec:body}
\begin{figure*}[tbp]
\centering
\includegraphics[width=\textwidth]{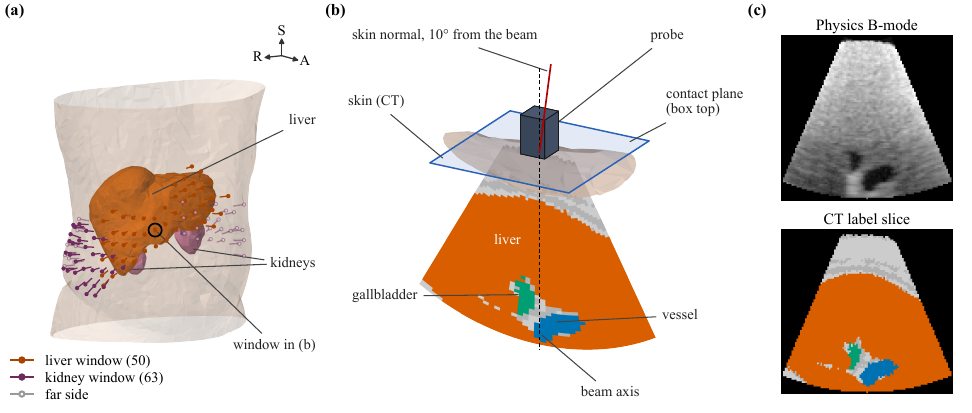}
\caption{The patient's CT skin, its screened windows and the contact plane, shown for training patient s0011.
(a)~The skin surface extracted from the patient's labeled CT, with the liver and kidneys inside.
Dots mark the faces of the screened acoustic windows on the skin: 50 for the liver and 63 for the kidneys.
Each stroke points along its window's beam into the body.
Hollow dots lie on the far side of the body from the viewer.
(b)~The probe seated at the top-ranked liver window, in the world frame of the MuJoCo scene.
The contact plane, the top face of the static MuJoCo box, passes through the window face perpendicular to the beam axis.
At this window the beam tilts $10^\circ$ from the skin normal toward the liver, so the plane is not tangent to the skin.
Contact uses only this plane, not the skin surface.
The fan is the slicing plane of the probe, painted with the CT labels it samples (orange liver, green gallbladder, blue vessel, grays other tissue).
(c)~The physics render at this pose and its CT label slice, mapped into the measured sector geometry without domain randomization and with a fixed speckle seed.}
\label{fig:scene}
\end{figure*}

Anatomy comes from 19 public TotalSegmentator CTs~\cite{totalseg}, split by patient into 13 training and six held-out cases that no component trained in this study saw (Fig.~\ref{fig:scene}).
The skin comes from the same CT, so probe seating needs no registration.
Each accepted window passes eight screens (Supplementary S5), and its beam follows the inward skin normal or tilts toward the organ by at most $10^\circ$.
Liver windows lie on the anterior to right anterolateral abdomen, kidney windows on the flanks, and some in-plane rotations break the clinical display convention (Supplementary S5).
No sonographer rated a window or its view.
Every held-out patient has at least 5 liver and 11 kidney windows.

USB Gym seats the probe at a window screened on the patient's CT skin.\phantomsection\label{sec:contact}
MuJoCo presses it against a plane through the target window face, perpendicular to the target beam.
Because the plane carries the target's depth and beam direction into the dynamics, we treat this target-anchored contact and the start lift onto it as privileged structure.
The plane is tangent to the skin only at 15 of the 60 evaluation windows and tilted from it by up to $10^\circ$ at the other 45.
The press is a position clamp at 5~mm below the plane, about 8.5~N in a steady press, and the tissue is rigid.
Each 0.1~s control step integrates the commanded twist into a target pose without feedback, then simulates 20~ms of physics through a weld with a 0.02~s time constant.
The probe is thus a lagged position servo.
Under the CT-mask servo base, it trails its command by a mean 3.40~mm with contact, press included, and 2.09~mm without (post hoc).
Starts below the plane, 40 of the 60, are lifted along its normal, the target beam axis, by a mean 6.4~mm (post hoc).
Afterwards, the axial start offset spans $-9.4$ to $-0.6$~mm, so no start lies deeper than the target face (post hoc).

\subsection{Rendering and the standard texture stage}\label{sec:rendering}
The reference renderer is a physics model seeded from the LOTUS defaults~\cite{velikova2023lotus} (every term, and what the model leaves out: Supplementary S1).
Each label slice is a $60^\circ$ fan of 160 scanlines with 512 samples to 120~mm depth.
Shadowing leaves at sample $s$ of scanline $l$ the energy
\begin{equation}
\begin{gathered}
E_{l,s} = \Bigl[\prod_{k=1}^{s} \left(1 - R_{l,k}\right)\Bigr] \exp\Bigl(-2\kappa\,\Delta z \sum_{j=1}^{s} \mu_{l,j}\Bigr),\\
R_{l,k} = \left(\frac{Z_{l,k} - Z_{l,k-1}}{Z_{l,k} + Z_{l,k-1}}\right)^{2},
\end{gathered} \label{eq:shadow}
\end{equation}
with $R_{l,1} = 0$, impedance $Z$, attenuation $\mu$, sample spacing $\Delta z$ and a fixed gain $\kappa$.
A depth-varying point spread function blurs a speckle field fixed to the image grid and drawn from one seed per run, or per episode under domain randomization (DR).
Targets and observations share this field, so they are identical images at the same pose.
The renderer's acoustic constants and DR ranges are heuristic, not fitted to measured images.
In the default render (default constants, no DR), organs look almost equally bright, that is, nearly isoechoic: their echo strengths differ by at most 1~dB.
Organ boundaries are not drawn either, because their echoes are too weak for the display's 50~dB range (Supplementary S1.10).
The standard texture stage, SonoSPADE, textures each organ from the CT label.
Frames are warped into the $46^\circ$ sector of the $128 \times 128$ Vitale frames~\cite{vitale}, which narrows the $60^\circ$ fan by a factor of 0.77.

SonoSPADE~\cite{sonospade}, the only stage reading the CT label, is the standard engine's texture stage, so label information reaches every standard frame (Section~\ref{sec:checks-label}). The pre-specified oracle ladder of the primary results also runs three other texture settings (Supplementary Fig.~\ref{fig:stages}): untextured physics, CUT~\cite{cut} and CycleGAN~\cite{cyclegan}.
We retrained the three networks once, with seed 0, on renders of eight training patients and 404 Vitale frames.
A learned mask comes from one of 12 texture-matched U-Net segmenters~\cite{ronneberger2015unet}, per stage and seed.
Each also trains on 42 annotated Vitale frames and the KidneyUS training split~\cite{kidneyus}, never on held-out patients, held-out real frames or Annotated Ultrasound Liver (AUL) frames~\cite{aul}.

\subsection{Match score and privileged channel}\label{sec:score}
The default reward, the match score $S$ of Eq.~\eqref{eq:score}, combines organ-mask overlap with an in-sector appearance term and uses no pose.
For frames $I$ and $I^{\ast}$ with organ masks $M$ and $M^{\ast}$,
\begin{equation}
\begin{gathered}
S = 0.6\, S_{\mathrm{seg}}(M, M^{\ast}) + 0.4\, S_{\mathrm{app}}(I, I^{\ast}),\\
S_{\mathrm{app}} = \tfrac{1}{2}\bigl(1 + \mathrm{NCC}_{\Omega}(I, I^{\ast})\bigr).
\end{gathered} \label{eq:score}
\end{equation}
Here $S_{\mathrm{seg}}$ is the mean per-organ intersection over union (IoU) of five organs, and $\mathrm{NCC}_{\Omega}$ is the normalized cross-correlation over the lit sector $\Omega$.
The step reward is the gain in $S$ minus a small action cost.
The score can therefore also rate real frames (Section~\ref{sec:results-real}), but its pose sensitivity on real images is untested.
The privileged channel contains the true pose error, both poses, the CT masks and the analytic twist (Box~\ref{box:reading}).
A controller meant for a physical rig must not read this channel, which the environment returns by default (Listing~\ref{lst:plug}).

\makeatletter
\@ifundefined{c@listing}{\floatstyle{plain}\newfloat{listing}{tbp}{lol}\floatname{listing}{Listing}\providecommand{\theHlisting}{\arabic{listing}}}{}
\makeatother

\section{Interface, plug protocol and compute}\label{sec:using}

\subsection{Observation, action and reward}\label{sec:interface}
A controller receives the observation of Section~\ref{sec:gym} and returns a normalized twist; each step's reward is the gain in $S$ minus the action cost (Section~\ref{sec:score}).
A texture stage, mask source or controller enters the environment through its texture or mask argument or its action, as in Listing~\ref{lst:plug}, and a CT case as a label volume with screened windows.
Plugged renderers and reward terms entered through adapters, code that replaces the render function or $S$ (Table~\ref{tab:slots}).
In the CT-free configuration, organ masks come from a learned segmenter, no pose, CT mask or privileged channel reaches the controller or the reward, and only evaluation code reads the pose error.
It still uses segmenters and a servo base fit with CT labels and true poses of training patients, and target-anchored contact (Section~\ref{sec:contact}).

\begin{listing*}[t]
\caption{One held-out evaluation episode with a toy texture stage, \texttt{my\_stage}, in the texture slot. Lines marked privileged use simulator state that no physical rig provides.}
\label{lst:plug}
\begin{lstlisting}[style=usbgpy]
import gymnasium as gym
import numpy as np
import usbg.gym_env  # registers the task

def my_stage(img01):  # texture slot: frame to frame
    return np.clip(img01, 0.0, 1.0) ** 1.5

# Building the environment calibrates the servo base of (*@{\color{black!60}Eq.~(\ref{eq:twist})}@*) on the 13 training patients.
env = gym.make("USBGym/ViewMatch-v0", base="servo",
    patients="test", texture=my_stage,  # slot call
    mask="ct",  # privileged
    privileged_info=True)  # privileged
obs, info = env.reset(options={"eval_episode": 0})  # patient s0358, liver target
for _ in range(12):
    twist = np.zeros(6, np.float32)  # action: zero residual
    obs, reward, term, trunc, info = env.step(twist)
# Under CT masks and a zero residual, the stage does not change the motion, so this prints
# about 16.62, the CT-mask servo base's final error (mm) in the run behind (*@{\color{black!60}Table~\ref{tab:ladder}}@*).
print(info["privileged"]["pose_error_mm"])  # privileged
\end{lstlisting}
\end{listing*}

\subsection{Plug protocol}\label{sec:plug-protocol}

Under the plug protocol, we swapped one slot at a time and held the rest of the loop at that slot's condition in Table~\ref{tab:pluggability} or Supplementary Table~\ref{tab:pluggability-engine}.
A new method therefore meets the same 60 held-out episodes as its slot's reference component and is evaluated at 12 steps only.
No plugged method is tuned: each runs with settings fixed before its held-out runs (controllers: Supplementary Table~S3.4).
Reinforcement learners get the reference controller's 57,600 environment steps, and imitation learners its 40,000 samples.
Learned methods run seeds 0 to 2, and SAM, MedSAM, SIMUS and the cross-entropy look-ahead run seed 0 only; plugged networks run batch normalization in evaluation mode.
Holm's adjustment runs within five pre-specified families of 4 texture stages, 7 mask sources, 10 controllers, 2 rewards and 2 renderers; five further families, specified after these (Section~\ref{sec:stats}), are never pooled with them.

\subsection{Compute cost}\label{sec:compute}

All compute costs were measured on one Apple M5 Pro laptop with 24~GB of memory (software versions: Supplementary Table~\ref{tab:compute}).
On its central processing unit (CPU), one closed-loop step with a learned-mask policy takes a median 41.5 to 103.4~ms over the four reference texture stages, or 9.7 to 24.1 steps per second.
Training one reference policy on the laptop's GPU took a median 0.79~h with CT masks, 1.02~h with learned masks and 1.32~h under domain randomization. All 27 reference training runs took 0.61 to 2.97~h, timed while three runs shared the laptop (every frame, step and training time: Supplementary Table~S3.3).
The testbed, the analysis scripts and the table and figure builds were written, tested and run with the AI coding assistant Claude Code (Anthropic), on the models Claude Opus 4.8, Opus 5 and Opus 5.5; those scripts generate every table and figure data file from the archived results.

\section{Reference pipeline and evaluation design}\label{sec:reference}

\subsection{Reference controller}\label{sec:controller}
The reference controller adds a bounded learned residual to the servo base:
\begin{equation}
a_t = \operatorname{clip}\bigl( b(M_t, M^{\ast}) + \rho_t,\; \pm a_{\max} \bigr), \label{eq:twist}
\end{equation}
where the residual $\rho_t$ is bounded at half the twist bound $a_{\max}$ (Section~\ref{sec:task}).
The servo base $b$ adapts the Mebarki et al.~\cite{mebarki2010} servo: least squares on 500 training frame pairs maps per-organ mask moments, area and centroid, linearly to the analytic twist under its own mask source.
The residual imitates the analytic twist minus the base, then trains by REINFORCE~\cite{reinforce}, a policy-gradient reinforcement learning (RL) method, with an imitation anchor; only training uses the true pose.
At inference the learned-mask servo reads no pose and no CT mask, a CT-free configuration evaluated only in simulation.
Policies keep batch normalization in training mode at evaluation (evaluation mode: Section~\ref{sec:checks-repeat}).

\subsection{Oracle ladder}\label{sec:ladder}
The oracle ladder's rungs see less and less simulator state on identical episodes, so the step between two rungs measures what that state is worth.
\begin{enumerate}[leftmargin=*,itemsep=0pt,topsep=2pt]
\item Pose oracle (privileged): the analytic twist.
\item CT-mask servo (privileged): Eq.~\eqref{eq:twist} with CT masks.
\item Learned-mask servo: Eq.~\eqref{eq:twist} with the stage's texture-matched segmenter.
\item Servo base alone: $b$ with no residual, privileged under CT masks.
\item Static: no commanded motion.
\item Scripted sweep: a fixed motion that reads no image and does not aim at the target. Over the 12 steps, the probe is pressed 1.2~mm into the body along the beam for four steps, then fanned about its lateral axis, 1.15$^{\circ}$ per step, four steps out and four back, each with a 0.4~mm press to keep contact (commanded values).
\end{enumerate}
Neither gap name of Box~\ref{box:reading} assigns a cause.
The RL contribution is a policy's final error minus its error after imitation.

\subsection{Episodes and metrics}\label{sec:eval}
Each configuration, or arm, runs 60 crossed episodes per seed, six held-out patients by two organs by five windows.
By design, every rung, seed and plugged method meets the same targets and starts.
Contact stays on and domain randomization (DR) off, except in the contact-off evaluations, one skin-tangent contact arm and one CT-mask DR ablation.

The primary metric is the final translation error in mm, the translational norm of the SE(3) logarithm of the final pose relative to the target.
Rotation error is the geodesic angle to the target orientation.
Success means ending within 5~mm and $5^\circ$, or within 10~mm and $10^\circ$, tolerances set here, not by a clinical source.
Post hoc secondary metrics are per-axis errors in the target probe frame and the keep-best error or IoU, the value at the highest-scoring step, selected without the pose.
Learnability is the Spearman correlation of the appearance term with closeness to the target pose.
Distances to real frames are the intensity Wasserstein-1 distance (W1), the kernel and Fr\'echet Inception distances (KID, FID)~\cite{kid,fid} and the area under the curve (AUC) of a classifier two-sample test (C2ST)~\cite{c2st}.

\subsection{Statistical analysis}\label{sec:stats}
A contrast is a per-patient paired difference of means over organ, window and seed.
Every contrast is conditional on the trained artifacts (texture networks, segmenters, bases and policies).
Primary results keep the pre-specified, unadjusted CI-and-sign rule of Box~\ref{box:reading}, on a 95\% patient-cluster bootstrap confidence interval (CI).
Its type-I error at $n = 6$ was 0.11 to 0.16 in simulation (post hoc).
Exploratory analyses therefore use the paired $t$ rule of Box~\ref{box:reading}, a two-sided paired $t$-test with 5 degrees of freedom, Holm-adjusted~\cite{holm1979} within each pre-specified family.
No exact test can reach Holm's threshold at $n = 6$ in a family of two or more, since the smallest two-sided exact sign-flip $p$ is $2/64$.
Exact sign-flip $p$ values and bootstrap CIs serve as sensitivity analyses.

The 20 replication CTs are screened in a seeded order, and a CT is accepted only with at least 2 kidney windows, the minimum that training and evaluation use.
Their decision rule is the exact sign-flip test with Holm.
An external replication cohort in simulation uses 30 CTs of the public AMOS 2022 dataset~\cite{ji2022amos}, with sign-flip $p$ values from 1,000,000 random sign vectors.
AMOS records no patient identifier, so the CT is the unit, and two CTs from one patient cannot be excluded.
The 79 eligible AMOS CTs are all 512 by 512 and 65 share one header pattern, so they may come from fewer scanners than AMOS as a whole; we make no statement about centers, vendors or scanners.

The pre-specified analyses were not formally pre-registered.
Results of an earlier version of this testbed on the same six held-out patients were known when the primary analyses were specified.
The plug protocol was fixed before any plugged method ran on held-out or cohort patients.
The paired $t$ rule was chosen after post hoc paired $t$-tests on the 25 ladder contrasts had been seen (Section~\ref{sec:ladder-contact}).
Some exploratory runs began before their decision rules were fixed (Supplementary Table~S4.1).
They are five of the gap-cause level sets, the seed-0 look-ahead runs, SonoSPADE's undirected-motion episodes, the planted-deficit RL runs and the two extra SonoSPADE texture seeds of the texture-budget test.
The SAMUS and LPIPS priors were set after the SAM, MedSAM and reward results had been seen.
Three further exploratory experiments were each fixed before their runs: learned-mask controllers with new learner seeds, skin-tangent contact and canonical phantoms.
Earlier learned-mask controller results have no pre-specified rule (\ref{app:decisions-earlier}).
A benchmark baseline is set on the 20 cohort CTs by a rule fixed before those runs, its components known.
A final set of follow-up runs was specified after nearly all other results had been seen, though before its own runs.
It has no decision rule, and each of its results is labeled estimation only (Supplementary S7).

\section{Results: the reference pipeline}\label{sec:results-ladder}

\subsection{Absolute performance and the oracle-to-CT-mask gap}\label{sec:ladder-absolute}

\begin{table*}[tp]
\caption{Oracle ladder: final translation error (mm), 6 held-out patients, contact on, evaluation domain randomization off; policy rows average 3 seeds, each meeting the same 60 crossed episodes. Pre-specified results, primary; the last block is post hoc. Errors: mean of per-patient means. Gaps: mean per-patient paired difference, patients of 6 sharing its sign, and 95\% patient-cluster bootstrap CI (10,000 resamples); $^{\ast}$a difference under the pre-specified CI-and-sign rule. Every contrast is conditional on the trained artifacts (texture networks, segmenters, bases and policies). Seeds index artifacts; the patient is the sampling unit. The servo base accounts for nearly all of the improvement; imitation and REINFORCE each change the final translation error by less than 0.5~mm. $^{\mathrm{L}}$Reads the CT label slice. $^{\dagger}$Privileged: simulator state that no rig provides. Shaded: SonoSPADE, the texture stage of the standard engine and the main benchmark; physics, CUT and CycleGAN are the ladder's other texture settings.}
\label{tab:ladder}
\centering
\renewcommand{\baselinestretch}{1}\footnotesize
\begin{tabular}{@{}l ccc>{\columncolor{black!8}}c@{}}
\toprule
 & Physics & CUT & CycleGAN & SonoSPADE$^{\mathrm{L}}$ \\
\midrule
\multicolumn{5}{@{}l}{\textit{Final translation error (mm)}} \\
Static start (no commanded motion) & \multicolumn{4}{c}{16.03} \\
Scripted sweep (no feedback) & \multicolumn{4}{c}{15.52} \\
Analytic pose oracle$^{\dagger}$ & \multicolumn{4}{c}{1.53} \\
\addlinespace
\multicolumn{5}{@{}l}{CT mask$^{\dagger}$} \\
\quad Servo base & 11.65 & 11.65 & 11.65 & 11.65 \\
\quad After imitation & 11.78 & 11.72 & 11.70 & 11.74 \\
\quad Final & 11.81 & 11.64 & 11.69 & 11.69 \\
\addlinespace
\multicolumn{5}{@{}l}{Learned mask (texture-matched segmenter)} \\
\quad Servo base & 17.20 & 16.84 & 16.85 & 13.13 \\
\quad After imitation & 17.02 & 16.42 & 16.42 & 13.07 \\
\quad Final & 16.73 & 16.90 & 16.50 & 13.11 \\
\midrule
\multicolumn{5}{@{}l}{\textit{Gaps (mm): patient-level mean, k/6 patients, 95\% CI}} \\
Oracle-to-CT-mask gap & \begin{tabular}[c]{@{}c@{}}$+10.28$$^{\ast}$, 6/6\\$[+9.21, +11.39]$\end{tabular} & \begin{tabular}[c]{@{}c@{}}$+10.11$$^{\ast}$, 6/6\\$[+8.96, +11.27]$\end{tabular} & \begin{tabular}[c]{@{}c@{}}$+10.17$$^{\ast}$, 6/6\\$[+9.06, +11.36]$\end{tabular} & \begin{tabular}[c]{@{}c@{}}$+10.16$$^{\ast}$, 6/6\\$[+9.10, +11.27]$\end{tabular} \\[2pt]
CT-to-learned-mask gap & \begin{tabular}[c]{@{}c@{}}$+4.92$$^{\ast}$, 5/6\\$[+2.46, +6.79]$\end{tabular} & \begin{tabular}[c]{@{}c@{}}$+5.26$$^{\ast}$, 6/6\\$[+3.07, +6.95]$\end{tabular} & \begin{tabular}[c]{@{}c@{}}$+4.80$$^{\ast}$, 5/6\\$[+2.15, +6.42]$\end{tabular} & \begin{tabular}[c]{@{}c@{}}$+1.43$$^{\ast}$, 6/6\\$[+0.86, +2.11]$\end{tabular} \\[2pt]
RL contribution, CT mask & \begin{tabular}[c]{@{}c@{}}$+0.04$, 3/6\\$[-0.10, +0.17]$\end{tabular} & \begin{tabular}[c]{@{}c@{}}$-0.08$, 4/6\\$[-0.29, +0.10]$\end{tabular} & \begin{tabular}[c]{@{}c@{}}$0.00$, 3/6\\$[-0.23, +0.21]$\end{tabular} & \begin{tabular}[c]{@{}c@{}}$-0.06$, 3/6\\$[-0.33, +0.22]$\end{tabular} \\[2pt]
RL contribution, learned mask & \begin{tabular}[c]{@{}c@{}}$-0.29$, 3/6\\$[-0.80, +0.17]$\end{tabular} & \begin{tabular}[c]{@{}c@{}}$+0.47$, 4/6\\$[+0.06, +0.90]$\end{tabular} & \begin{tabular}[c]{@{}c@{}}$+0.08$, 3/6\\$[-0.37, +0.53]$\end{tabular} & \begin{tabular}[c]{@{}c@{}}$+0.05$, 3/6\\$[-0.24, +0.33]$\end{tabular} \\[2pt]
SonoSPADE minus stage, learned-mask final & \begin{tabular}[c]{@{}c@{}}$-3.62$$^{\ast}$, 5/6\\$[-5.03, -1.67]$\end{tabular} & \begin{tabular}[c]{@{}c@{}}$-3.78$$^{\ast}$, 5/6\\$[-5.14, -2.05]$\end{tabular} & \begin{tabular}[c]{@{}c@{}}$-3.38$$^{\ast}$, 5/6\\$[-4.94, -1.19]$\end{tabular} & n/a \\
\midrule
\multicolumn{5}{@{}l}{\textit{Post hoc block}} \\
Final rotation error (deg) & \multicolumn{4}{l}{static 13.70, scripted 5.63, pose oracle$^{\dagger}$ 1.02} \\
\quad CT mask$^{\dagger}$, final & 7.60 & 7.79 & 7.35 & 6.91 \\
\quad Learned mask, final & 7.93 & 8.29 & 7.10 & 8.26 \\
Success, 5~mm and 5$^{\circ}$ (\%) & \multicolumn{4}{l}{static 0.0, scripted 3.3, pose oracle$^{\dagger}$ 100.0} \\
\quad CT mask$^{\dagger}$, final & 8.3 & 8.9 & 9.4 & 10.0 \\
\quad Learned mask, final & 1.1 & 3.9 & 2.2 & 5.6 \\
Success, 10~mm and 10$^{\circ}$ (\%) & \multicolumn{4}{l}{static 3.3, scripted 16.7, pose oracle$^{\dagger}$ 100.0} \\
\quad CT mask$^{\dagger}$, final & 32.8 & 33.9 & 38.3 & 36.7 \\
\quad Learned mask, final & 21.1 & 14.4 & 18.9 & 30.6 \\
Final match score $S$ & \multicolumn{4}{l}{static 0.462, scripted 0.515, pose oracle$^{\dagger}$ 0.872} \\
\quad CT mask$^{\dagger}$, final & 0.610 & 0.616 & 0.612 & 0.624 \\
\quad Learned mask, final & 0.472 & 0.472 & 0.483 & 0.555 \\
\bottomrule
\end{tabular}
\end{table*}

From a 16.03~mm static start, the privileged pose oracle nearly reaches the target, at 1.53~mm, but the CT-mask servo stays far from it, at 11.64 to 11.81~mm (primary; Table~\ref{tab:ladder}).
The oracle-to-CT-mask gap is $+10.28$~mm [$+9.21$, $+11.39$] with physics frames (6/6) and within 0.17~mm of that with the other stages.

Without CT masks, three of the four learned-mask servos end at 16.50 to 16.90~mm, $+0.5$ to $+0.9$~mm above the static start (paired $t$ CIs $-2.00$ to $+3.41$~mm; post hoc; Supplementary Fig.~\ref{fig:ladder}).
SonoSPADE's ends at 13.11~mm, $-2.92$~mm [$-5.73$, $-0.10$] from the static start, 5/6 (paired $t$ $p$ 0.045, not detectable after Holm; post hoc).
At 5~mm and 5$^{\circ}$, 8.3 to 10.0\% of CT-mask and 1.1 to 5.6\% of learned-mask final episodes succeed, against 100\% for the oracle (post hoc).

\subsection{The CT-to-learned-mask gap by texture stage}\label{sec:ladder-gap}

The CT-to-learned-mask gap is larger with physics, CUT and CycleGAN frames, $+4.92$, $+5.26$ and $+4.80$~mm (5/6, 6/6 and 5/6), than with SonoSPADE frames, $+1.43$~mm [$+0.86$, $+2.11$] (6/6).
The gap compares policies trained under different mask sources, so it does not estimate what better masks would remove.
With texture-matched segmenters, per-tissue texture gives the lowest learned-mask error.
Its learned-mask final lies 3.38 to 3.78~mm below the others' (5/6 each).
SonoSPADE alone also reads the CT label slice (Section~\ref{sec:checks-label}).
Per-tissue texture gives the most pose-sensitive appearance term of the four stages.
Its learnability is 0.613, against at most 0.548 for the others (Supplementary Table~\ref{tab:learnability}).

\subsection{Fine-tuning and translation error}\label{sec:ladder-finetune}

The servo base accounts for nearly all of the improvement; imitation and REINFORCE each change the final translation error by less than 0.5~mm.
The CT-mask servo base alone ends at 11.65~mm.
The largest RL contribution, $+0.47$~mm [$+0.06$, $+0.90$] with CUT frames and learned masks, falls short of the decision rule (4/6).

\subsection{Contact, texture under CT masks and domain randomization}\label{sec:ladder-contact}

With contact off at evaluation, the CT-mask servo ends at 14.4 to 14.8~mm, 2.7 to 3.0~mm farther (6/6 per stage).
With contact, the scripted sweep, which reads no image, also ends with smaller tilts: 3.57 and 3.10$^\circ$ out of plane and in plane, no larger at tilted windows, against 9.67 and 8.30$^\circ$ without contact (post hoc).
Every servo in Table~\ref{tab:ladder} thus owes part of its absolute accuracy to the target-anchored plane (Section~\ref{sec:contact}).
On 20 new CTs, contact off raises the final error by 2.14~mm [1.68, 2.60] (20/20) for the CT-mask servo base and 2.02 to 2.13~mm for CT-mask final policies (exploratory; estimation only; Supplementary S7).
With contact on and CT masks, only physics against SonoSPADE frames has a stage-pair CI excluding 0, $+0.13$~mm [$+0.02$, $+0.25$], 4/6, short of the decision rule.
No interval of the domain-randomization ablation excludes zero (descriptive; \ref{app:dr} and~\ref{app:contact-dr}).

Re-tested post hoc with paired $t$-tests, Holm-adjusted within each family, 11 of the 25 ladder contrasts reach Holm's threshold: every oracle-to-CT-mask gap, every CT-to-learned-mask gap and every learned-mask SonoSPADE-against-stage contrast.
On the learnability, realism, domain-randomization policy and contact contrasts of the primary analysis, the paired $t$-test reaches Holm's threshold in 9 of 9, 1 of 1, 0 of 3 and 9 of 16 (post hoc). The exact test reaches Holm's threshold only in the one-member realism family, at $p = 2/64$ (post hoc; Supplementary Table~S2.18).

\subsection{What limits the CT-mask servo}\label{sec:gap-cause}

The CT-mask servo leaves headroom in the match score.
This subsection is exploratory.
Privileged look-ahead values average look-ahead seeds 0 to 2.
The servo base barely predicts the analytic twist on five of six axes (held-out $R^2$ at most 0.13).
Under CT masks, its elevational error grows from 8.08~mm at step 12 to 13.75~mm at step 36, and its roll from 4.14 to 10.68$^{\circ}$ (6/6 each; post hoc).
The oracle-to-CT-mask gap is not attributed to one cause (conditional on this renderer: grid-fixed speckle, no elevational beam width, rigid tissue).
Under CT masks, 0/60 (0\%) of targets have a pose 5~mm or more away scoring within 0.01 of the target's own score (0/6 patients with a majority).
Privileged look-ahead ascent on the CT-mask match score ends at 8.55~mm, against 11.65~mm for the servo base and 1.53~mm for the pose oracle, that is, 3.10~mm [0.14, 6.05] below the servo base (6/6; conditional on this renderer).
It meets the paired $t$ rule ($t$ $p$ = Holm $p$ = 0.043; exact sign-flip $p$ = 0.031).
Pose-free hill-climbing keeps its best random step on $S$.
Under CT masks, hill-climbing on the match score ends 4.03~mm above the servo base (5/6).
It also meets the paired $t$ rule ($t$ CI [$+0.93$, $+7.13$]~mm; $t$ $p$ 0.021; Holm $p$ 0.041; exact sign-flip $p$ 0.063).
The testbed exposes headroom between the servo base (11.65~mm) and the pose oracle (1.53~mm) that this controller does not close.

\section{Results: published methods in the guide's three slots}\label{sec:pluggability}

Of the 19 contrasts in the guide's three slot families, 4 differ and 15 do not, 8 of these under privileged CT masks and 2 within $\pm 1$~mm (post hoc TOST; Table~\ref{tab:pluggability}, Supplementary Fig.~\ref{fig:forest}).
Post hoc slot MDDs are 0.085 to 2.14~mm, smallest in the reward slot, where images act only through the residual.
\label{sec:plug-coverage}All 27 published methods ran under one protocol (Section~\ref{sec:plug-protocol}) with pre-specified rules (Section~\ref{sec:stats}); the results are exploratory, and \ref{app:decisions} gives each decision. This section reports the 20 in the guide's slots; the 7 in the engine's slots are in Supplementary Table~\ref{tab:pluggability-engine}.

\begin{table*}[tp]
\renewcommand{\baselinestretch}{1}\scriptsize\setlength{\fboxsep}{0.6pt}
\caption{Published methods in the three slots of the guide: mask source, controller and reward (exploratory, pre-specified rule). Every row: 12 steps, the 60 held-out episodes of Table~\ref{tab:ladder}, contact on. Error: final translation error, mean over seeds of patient means. Difference: from the slot's reference component, paired by seed, with 95\% paired $t$ CI and patients of 6 sharing its sign. Seeds: the seeds, of 3, whose own difference shares that sign (post hoc; n/a for seed-0 rows). Holm $p$: within the slot's family. TOST: post hoc and unadjusted, yes when the 90\% paired $t$ CI lies inside $\pm 1$~mm, the pre-specified equivalence margin. Decision: the pre-specified outcome words; $^{\mathrm{e}}$equivalence within $\pm 1$~mm not shown. A decision of no detectable difference (in the pretrained-backbone family, does not detectably change) is not powered for equivalence; the TOST column shows where equivalence within $\pm 1$~mm holds. $^{\mathrm{a}}$Unadjusted CI excludes 0; not declared a difference after Holm (post hoc note). Slot headers give the condition, the reference error and the post hoc minimum detectable difference (MDD) from the median SD of the slot's contrasts with its reference, over the rows carried to the cohort analysis, so not SAM, MedSAM, CEM or the CT label, also at the first Holm level. The MDD is that of the paired $t$-test alone; the 5 of 6 sign condition makes the decision rule slightly less sensitive. Every contrast is conditional on the trained artifacts (texture networks, segmenters, bases and policies). Seeds index artifacts; the patient is the sampling unit. Costs, $t$ and sign-flip $p$, the 90\% CIs and scores are in the supplementary twin of this table. $^{\mathrm{L}}$Reads the CT label slice. $^{\dagger}$Privileged: simulator state that no rig provides. $^{\ddagger}$Fewer seeds than designed.}
\label{tab:pluggability}
\centering
\setlength{\tabcolsep}{2.4pt}
\begin{tabularx}{\textwidth}{@{}>{\raggedright\arraybackslash}X r l c r c >{\raggedright\arraybackslash}p{0.2\textwidth}@{}}
\toprule
Slot and method & \begin{tabular}[b]{@{}r@{}}Error\\(mm)\end{tabular} & Difference (mm), 95\% $t$ CI, $k$/6 & Seeds & Holm $p$ & TOST & Decision \\
\midrule
\multicolumn{7}{@{}>{\raggedright\arraybackslash}p{0.985\textwidth}@{}}{\textbf{Mask source}. Condition: SonoSPADE frames; mask from the plugged source; servo base fitted under that mask, residual off; seeds 0 to 2, and seed 0 for the SAM-type rows, paired with seed 0. Reference: the stage-matched U-Net, 13.13~mm. Families: mask source ($m = 7$), second mask source ($m = 5$) and pretrained backbone ($m = 4$), never pooled. Post hoc MDD: 2.14~mm, 3.25~mm at $\alpha = 0.05/7$.} \\
\multicolumn{7}{@{}l}{\quad\textit{Reference}} \\
\quad U-Net, stage-matched (authors')~\cite{ronneberger2015unet} & 13.13 &  &  &  &  &  \\
\quad\quad Seed 0 only, for the seed-0 rows & 13.33 &  &  &  &  &  \\
\multicolumn{7}{@{}l}{\quad\textit{Family ($m = 7$)}} \\
\quad LR-ASPP, MobileNetV3-L~\cite{howard2019mobilenetv3} & 12.76 & $-0.38$ $[-2.01, +1.25]$, 4/6 & 2/3 & $1.000$ & no & no detectable difference \\
\quad DeepLabV3, MobileNetV3-L~\cite{chen2017deeplabv3,howard2019mobilenetv3} & 14.54 & $+1.41$ $[+0.15, +2.67]$, 6/6 & 3/3 & $0.242$ & no & no detectable difference$^{\mathrm{a}}$ \\
\quad DeepLabV3, ResNet-50~\cite{chen2017deeplabv3,he2016resnet} & 13.20 & $+0.06$ $[-1.51, +1.64]$, 3/6 & 2/3 & $1.000$ & no & no detectable difference \\
\quad FCN, ResNet-50~\cite{long2015fcn,he2016resnet} & 12.81 & $-0.32$ $[-1.58, +0.93]$, 4/6 & 1/3 & $1.000$ & no & no detectable difference \\
\quad nnU-Net 2D, 2 epochs~\cite{isensee2021nnunet} & 14.88 & $+1.74$ $[-0.20, +3.69]$, 4/6 & 3/3 & $0.416$ & no & no detectable difference \\
\quad SAM, boxes from the U-Net mask~\cite{kirillov2023sam} & 14.32 & $+0.99$ $[-0.65, +2.64]$, 5/6 & n/a & $0.904$ & no & no detectable difference \\
\quad MedSAM, boxes from the U-Net mask~\cite{ma2024medsam} & 12.85 & $-0.47$ $[-2.74, +1.79]$, 5/6 & n/a & $1.000$ & no & no detectable difference \\
\multicolumn{7}{@{}l}{\quad\textit{Second mask-source family ($m = 5$)}} \\
\quad SAMUS, boxes from the U-Net mask~\cite{lin2024samus} & 14.29 & $+0.96$ $[-0.37, +2.30]$, 5/6 & n/a & $0.613$ & no & no detectable difference \\
\quad LR-ASPP, MobileNetV3-L, ImageNet backbone~\cite{howard2019mobilenetv3} & 13.25 & $+0.11$ $[-1.22, +1.45]$, 4/6 & 2/3 & $1.000$ & no & no detectable difference \\
\quad DeepLabV3, MobileNetV3-L, ImageNet backbone~\cite{chen2017deeplabv3,howard2019mobilenetv3} & 12.66 & $-0.47$ $[-2.04, +1.09]$, 3/6 & 2/3 & $1.000$ & no & no detectable difference \\
\quad DeepLabV3, ResNet-50, ImageNet backbone~\cite{chen2017deeplabv3,he2016resnet} & 12.62 & $-0.52$ $[-1.92, +0.89]$, 4/6 & 2/3 & $1.000$ & no & no detectable difference \\
\quad FCN, ResNet-50, ImageNet backbone~\cite{long2015fcn,he2016resnet} & 12.77 & $-0.36$ $[-1.16, +0.44]$, 4/6 & 2/3 & $1.000$ & yes & no detectable difference \\
\multicolumn{7}{@{}l}{\quad\textit{Pretrained-backbone family, minus the same network from scratch ($m = 4$)}} \\
\quad LR-ASPP, MobileNetV3-L, ImageNet backbone~\cite{howard2019mobilenetv3}, minus the same network trained from scratch &  & $+0.49$ $[-1.92, +2.91]$, 3/6 & 3/3 & $1.000$ & no & does not detectably change \\
\quad DeepLabV3, MobileNetV3-L, ImageNet backbone~\cite{chen2017deeplabv3,howard2019mobilenetv3}, minus the same network trained from scratch &  & $-1.88$ $[-4.45, +0.69]$, 4/6 & 3/3 & $0.473$ & no & does not detectably change \\
\quad DeepLabV3, ResNet-50, ImageNet backbone~\cite{chen2017deeplabv3,he2016resnet}, minus the same network trained from scratch &  & $-0.58$ $[-2.33, +1.16]$, 4/6 & 3/3 & $1.000$ & no & does not detectably change \\
\quad FCN, ResNet-50, ImageNet backbone~\cite{long2015fcn,he2016resnet}, minus the same network trained from scratch &  & $-0.04$ $[-1.50, +1.42]$, 2/6 & 2/3 & $1.000$ & no & does not detectably change \\
\multicolumn{7}{@{}l}{\quad\textit{Secondary}} \\
\quad nnU-Net 2D, 20 epochs~\cite{isensee2021nnunet}$^{\ddagger}$ & 12.76 &  &  &  &  & secondary \\
\quad SAMUS, boxes from the U-Net mask~\cite{lin2024samus}, seeds 0 to 2 against U-Net seeds 0 to 2 & 13.98 & $+0.85$ $[-0.04, +1.74]$, 4/6 & 3/3 &  & no & secondary \\
\multicolumn{7}{@{}l}{\quad\textit{Privileged, secondary}} \\
\quad CT label (authors')$^{\dagger}$ & 11.65 & $-1.48$ $[-2.11, -0.86]$, 6/6 & 3/3 &  & no & secondary \\
\quad SAM, boxes from the CT label~\cite{kirillov2023sam}$^{\dagger}$ & 12.60 & $-0.73$ $[-3.36, +1.91]$, 3/6 & n/a &  & no & secondary \\
\quad MedSAM, boxes from the CT label~\cite{ma2024medsam}$^{\dagger}$ & 12.12 & $-1.21$ $[-3.76, +1.35]$, 4/6 & n/a &  & no & secondary \\
\quad SAMUS, boxes from the CT label~\cite{lin2024samus}$^{\dagger}$ & 13.52 & $+0.20$ $[-1.81, +2.20]$, 3/6 & n/a &  & no & secondary \\
\quad SAMUS, boxes from the CT label~\cite{lin2024samus}, seeds 0 to 2 against U-Net seeds 0 to 2$^{\dagger}$ & 13.48 & $+0.35$ $[-0.37, +1.07]$, 3/6 & 2/3 &  & yes & secondary \\
\bottomrule
\end{tabularx}
\end{table*}

\begin{table*}[tp]
\renewcommand{\baselinestretch}{1}\scriptsize\setlength{\fboxsep}{0.6pt}
\ContinuedFloat
\caption[]{Continued.}
\centering
\setlength{\tabcolsep}{2.4pt}
\begin{tabularx}{\textwidth}{@{}>{\raggedright\arraybackslash}X r l c r c >{\raggedright\arraybackslash}p{0.2\textwidth}@{}}
\toprule
Slot and method & \begin{tabular}[b]{@{}r@{}}Error\\(mm)\end{tabular} & Difference (mm), 95\% $t$ CI, $k$/6 & Seeds & Holm $p$ & TOST & Decision \\
\midrule
\multicolumn{7}{@{}>{\raggedright\arraybackslash}p{0.985\textwidth}@{}}{\textbf{Controller}. Condition: SonoSPADE frames; CT mask$^{\dagger}$; the plugged controller, residual or full twist as named; seeds 0 to 2, and seed 0 for CEM. Reference: the REINFORCE residual from the imitation start, 11.69~mm. Family: $m = 10$. Post hoc MDD: 2.09~mm, 3.41~mm at $\alpha = 0.05/10$.} \\
\multicolumn{7}{@{}l}{\quad\textit{Reference}} \\
\quad REINFORCE residual, imitation start (authors')~\cite{reinforce} & 11.69 &  &  &  &  &  \\
\quad\quad Seed 0 only, for the seed-0 rows & 11.79 &  &  &  &  &  \\
\multicolumn{7}{@{}l}{\quad\textit{Family ($m = 10$)}} \\
\quad PPO, residual~\cite{schulman2017ppo,sb3} & 11.50 & $-0.19$ $[-0.53, +0.16]$, 4/6 & 2/3 & $0.885$ & yes & no detectable difference \\
\quad SAC, residual~\cite{haarnoja2018sac,sb3} & 12.06 & $+0.37$ $[-0.86, +1.61]$, 4/6 & 1/3 & $0.925$ & no & no detectable difference \\
\quad TD3, residual~\cite{fujimoto2018td3,sb3} & 20.22 & $+8.54$ $[+7.00, +10.08]$, 6/6 & 3/3 & $<0.001$ & no & higher \\
\quad A2C, residual~\cite{mnih2016a3c,sb3} & 18.43 & $+6.74$ $[+5.41, +8.07]$, 6/6 & 3/3 & $<0.001$ & no & higher \\
\quad PPO, full twist~\cite{schulman2017ppo,sb3} & 14.34 & $+2.66$ $[-0.39, +5.70]$, 5/6 & 3/3 & $0.376$ & no & no detectable difference \\
\quad DQN, 13 discrete moves~\cite{mnih2015dqn,sb3} & 16.41 & $+4.72$ $[+1.64, +7.81]$, 6/6 & 3/3 & $0.077$ & no & no detectable difference$^{\mathrm{a}}$ \\
\quad BC, residual~\cite{pomerleau1991} & 11.25 & $-0.44$ $[-1.42, +0.55]$, 4/6 & 3/3 & $0.925$ & no & no detectable difference \\
\quad DAgger, residual~\cite{ross2011dagger} & 10.85 & $-0.84$ $[-3.42, +1.74]$, 4/6 & 2/3 & $0.925$ & no & no detectable difference \\
\quad BC, full twist~\cite{pomerleau1991} & 19.60 & $+7.92$ $[+5.10, +10.73]$, 6/6 & 3/3 & $0.006$ & no & higher \\
\quad DAgger, full twist~\cite{ross2011dagger} & 17.66 & $+5.98$ $[+0.53, +11.42]$, 5/6 & 3/3 & $0.222$ & no & no detectable difference$^{\mathrm{a}}$ \\
\multicolumn{7}{@{}l}{\quad\textit{Secondary}} \\
\quad Servo base alone (authors') & 11.65 & $-0.04$ $[-0.36, +0.29]$, 3/6 & 2/3 &  & yes & secondary \\
\quad REINFORCE residual, BatchNorm in eval mode (authors') & 11.70 & $+0.01$ $[-0.28, +0.30]$, 3/6 & 1/3 &  & yes & secondary \\
\quad BC, residual~\cite{pomerleau1991}, BatchNorm in train mode & 12.67 & $+0.98$ $[-0.02, +1.99]$, 5/6 & 3/3 &  & no & secondary \\
\quad DAgger, residual~\cite{ross2011dagger}, BatchNorm in train mode & 12.89 & $+1.20$ $[+0.57, +1.83]$, 6/6 & 3/3 &  & no & secondary \\
\quad BC, full twist~\cite{pomerleau1991}, BatchNorm in train mode & 15.75 & $+4.07$ $[+0.94, +7.19]$, 5/6 & 3/3 &  & no & secondary \\
\quad DAgger, full twist~\cite{ross2011dagger}, BatchNorm in train mode & 15.15 & $+3.47$ $[+0.83, +6.10]$, 5/6 & 3/3 &  & no & secondary \\
\multicolumn{7}{@{}l}{\quad\textit{Privileged, secondary}} \\
\quad CEM look-ahead~\cite{rubinstein1999cem}$^{\dagger}$ & 6.98 & $-4.81$ $[-9.02, -0.60]$, 5/6 & n/a &  & no & secondary \\
\addlinespace[4pt]
\multicolumn{7}{@{}>{\raggedright\arraybackslash}p{0.985\textwidth}@{}}{\textbf{Reward}. Condition: SonoSPADE frames; CT mask$^{\dagger}$; the REINFORCE residual from the imitation start, trained with the plugged potential; seeds 0 to 2. Reference: the match score $S$, retrained$^{\mathrm{b}}$, 11.97~mm. Families: reward ($m = 2$) and LPIPS reward ($m = 1$), never pooled. Post hoc MDD: 0.085~mm, 0.099~mm at $\alpha = 0.05/2$.} \\
\multicolumn{7}{@{}l}{\quad\textit{Reference}} \\
\quad Match score $S$ (authors') & 11.97 &  &  &  &  &  \\
\multicolumn{7}{@{}l}{\quad\textit{Family ($m = 2$)}} \\
\quad SSIM appearance term~\cite{wang2004ssim} & 11.92 & $-0.047$ $[-0.085, -0.009]$, 5/6 & 2/3 & $0.049$ & yes & lower \\
\quad NMI appearance term~\cite{studholme1999nmi} & 11.98 & $+0.006$ $[-0.042, +0.053]$, 4/6 & 1/3 & $0.774$ & yes & no detectable difference \\
\multicolumn{7}{@{}l}{\quad\textit{LPIPS reward family ($m = 1$)}} \\
\quad LPIPS appearance term~\cite{zhang2018lpips} & 11.90 & $-0.066$ $[-0.122, -0.011]$, 5/6 & 2/3 & $0.028$ & yes & lower \\
\multicolumn{7}{@{}l}{\quad\textit{Secondary}} \\
\quad Organ overlap only (authors') & 11.91 & $-0.065$ $[-0.128, -0.003]$, 5/6 & 2/3 &  & yes & secondary \\
\quad Appearance NCC only (authors') & 12.06 & $+0.094$ $[-0.004, +0.191]$, 5/6 & 2/3 &  & yes & secondary \\
\quad NCC over the whole sector (authors') & 11.99 & $+0.023$ $[-0.004, +0.050]$, 5/6 & 3/3 &  & yes & secondary \\
\multicolumn{7}{@{}l}{\quad\textit{Privileged, secondary}} \\
\quad Change in pose distance~\cite{sonogym,hase2020}$^{\dagger}$ & 11.92 & $-0.049$ $[-0.180, +0.082]$, 4/6 & 2/3 &  & yes & secondary \\
\quad Pose distance plus organ area~\cite{bi2022vesnet}$^{\dagger}$ & 11.90 & $-0.070$ $[-0.158, +0.018]$, 5/6 & 3/3 &  & yes & secondary \\
\multicolumn{7}{@{}>{\raggedright\arraybackslash}p{0.985\textwidth}@{}}{\quad (b) The reference is the match score $S$, retrained in the same way as the plugged rewards. It ends 0.28~mm above its 11.69~mm in Table~\ref{tab:ladder}, more than the 0.17~mm spread across texture stages under CT masks.} \\
\bottomrule
\end{tabularx}
\end{table*}

\subsection{Mask source}\label{sec:plug-mask}
None of the first family's seven mask sources differs detectably from the stage-matched U-Net: they end at 12.76 to 14.88~mm on SonoSPADE frames, against 13.13~mm.
SonoSPADE frames give the smallest CT-to-learned-mask gap, so differences in this slot may be compressed.
The four torchvision networks~\cite{howard2019mobilenetv3,chen2017deeplabv3,long2015fcn,he2016resnet} train from random weights and nnU-Net~\cite{isensee2021nnunet} trains 2 epochs against its default of 1,000, all on the U-Net's simulated and real frames.

Neither SAMUS~\cite{lin2024samus} nor any of the four networks with ImageNet-pretrained backbones differs detectably from the U-Net.
Within the families, only FCN ResNet-50 with an ImageNet backbone is shown equivalent within $\pm 1$~mm (post hoc TOST).
SAMUS is prompted with boxes from the U-Net's own mask, not with its native point prompts, so this row measures what SAMUS adds to the U-Net's detections.
The pretrained networks use the from-scratch recipe (4,000 iterations), which is not tuned for pretrained backbones.
ImageNet pretraining is not shown to change the mask-slot results of the torchvision segmenters ($n = 6$).

\subsection{Controller}\label{sec:plug-ctrl}
Under privileged CT masks on SonoSPADE frames, none of the PPO~\cite{schulman2017ppo}, SAC~\cite{haarnoja2018sac}, behavior cloning (BC)~\cite{pomerleau1991} and DAgger~\cite{ross2011dagger} residuals differs detectably from the slot's reference, the REINFORCE residual~\cite{reinforce} at 11.69~mm.
Only PPO is shown equivalent within $\pm 1$~mm (post hoc TOST).
The TD3~\cite{fujimoto2018td3} and A2C~\cite{mnih2016a3c} residuals raise the error, by $+8.54$ and $+6.74$~mm (6/6 each; Holm $p < 0.001$).
Each residual adds at most $a_{\max}/2$ to the servo base's twist in Eq.~\eqref{eq:twist}, a bounded search around the base.
With seeds 0 and 1, TD3 moves the probe identically on SonoSPADE and physics frames in all 60 episodes, and each TD3 actor was found saturated, adding one constant offset at the residual bound whatever the image (post hoc).
Without the servo base, full-twist BC raises the error by $+7.92$~mm (6/6, Holm $p$ 0.006).
The other controllers without the base, full-twist PPO and DAgger and DQN~\cite{mnih2015dqn}, show no detectable difference.
The unadjusted intervals of full-twist DAgger and DQN exclude zero.
TD3, A2C, DQN and full-twist BC and DAgger all end above the 16.03~mm static start.
The Stable-Baselines3 (SB3)~\cite{sb3} learners start from random weights and REINFORCE from imitation; a decision describes this training (Section~\ref{sec:plug-protocol}), not the algorithm.

Without CT masks, one published controller detectably lowers its servo base's error, with physics frames only.
In a pre-specified learned-mask controller experiment, we retrained the DAgger and BC residuals with new learner seeds; at run time they read only the frame history and target frame (\ref{app:decisions-arma}).
They learned from a privileged pose-oracle teacher, on a base and segmenters fit with true poses and CT labels of training patients.
They were chosen by CT-mask results on the same six patients, so the experiment is not independent of their earlier runs; new seeds remove only the seed part of that selection.
Under learned masks with the physics stage, the DAgger residual gives a lower final translation error than the learned-mask servo base ($-3.17$~mm [$-5.04$, $-1.31$], 6/6; exploratory).
With physics frames and learned masks, the testbed separates 1 of the three method pairs of this arm at $n = 6$ (exploratory).
The separated pair is the DAgger residual and its base, which end at 14.02 and 17.20~mm, against 16.03~mm for the static start; DAgger and BC are not separated.
With SonoSPADE frames and learned masks, no pair of methods in this arm is separated at $n = 6$ (exploratory).
Equivalence within $\pm 1$~mm is not shown for any of its contrasts.
With physics frames, the BC residual's unadjusted interval excludes zero (Holm $p$ 0.177).

\subsection{Reward}\label{sec:plug-reward}
SSIM and LPIPS end below the retrained match score $S$, each by under a quarter of $S$'s 0.28~mm drift.
A new term replaces $S$ only in the REINFORCE step reward, under privileged CT masks, against $S$ retrained at 11.97~mm.
NMI~\cite{studholme1999nmi} shows no detectable difference.
A reward row acts only through REINFORCE, which moved the error of the primary results by under 0.5~mm and did not recover a planted deficit, so a null row says little.
The SSIM~\cite{wang2004ssim} appearance term gives a lower final translation error than the paper's $S$ in the reward potential slot ($-0.047$~mm $[-0.085, -0.009]$, 5/6; exploratory).
The LPIPS~\cite{zhang2018lpips} appearance term gives a lower final translation error than the paper's $S$ in the reward slot ($-0.066$~mm $[-0.122, -0.011]$, 5/6; exploratory).
Both meet the paired $t$ rule (Holm $p$ 0.049 and 0.028; exact sign-flip $p$ 0.0625 each).
For both, the pre-specified patient-by-seed interval excludes 0, while a post hoc crossed one includes it.
Every plugged reward is equivalent to $S$ within $\pm 1$~mm (post hoc TOST).
The appearance terms train only with target and observation sharing one speckle field.

\subsection{Benchmark baseline}\label{sec:plug-best}

To test the testbed, we combined slot components: the authors' unless a slot decision found a method lower.
Fixed before its runs, its components known, the rule used the six held-out patients, not the 20 cohort CTs that test it (Section~\ref{sec:checks-cohort}).
Only the reward slot changed, to the SSIM appearance term, so no pipeline plugged in several slots was measured.
That choice rests on a small difference: SSIM's $-0.047$~mm under privileged CT masks with SonoSPADE frames lies below the slot's post hoc MDD of 0.085~mm. Its paired $t$ $p$ of 0.0246 lies 0.0004 below Holm's first threshold, and its sign held in 2 of 3 learner seeds, shared with the retrained $S$.
With SSIM, the residual ends at 11.92~mm, above the primary REINFORCE result of 11.69~mm.
The combination's imitation start, behavior-cloning anchor, servo base and segmenters were fit with true poses and CT labels of training patients.

This test runs on the standard engine; its physics-frame arm is reported in \ref{app:decisions-combination}.
On 20 new CTs, the SonoSPADE combination gives a lower final translation error than the authors' original pipeline ($-0.26$~mm [$-0.39$, $-0.12$], 16/20, Holm $p = 0.002$; exploratory).
So, at 13.70~mm [12.68, 14.73], it is the benchmark baseline on the standard engine (exploratory), whose SonoSPADE frames are rendered from the CT label slice, so label information reaches the image.
It is equivalent to its learned-mask servo base within $\pm 1$~mm ($-0.01$~mm; TOST, Holm $p = 3.8 \times 10^{-6}$).
It is also equivalent to our reference pipeline (Holm $p = 3.8 \times 10^{-6}$).
Its difference from it is conditional on its six RL runs, and its crossed case-by-seed interval includes 0, so it may come from those runs rather than from the change of reward.
Retrained with the same procedure and seeds, $S$ ends 0.04~mm below the combination and differs from our reference pipeline by $-0.29$~mm (estimation only).
This control holds the training procedure and the seeds fixed; it does not make a difference of the combination a property of SSIM in general.
The combination ends far above the privileged pose oracle, by 12.24~mm (20/20; Holm $p = 3.8 \times 10^{-6}$; conditional on this renderer).
The rule (Supplementary Table~\ref{tab:combination}) ranks no other pipeline, such as the scripted sweep, the learned-mask controller arm or those of Supplementary S7.

A post hoc selection (Supplementary Table~\ref{tab:best-so-far}), whose outcome predates the pre-specified rule, takes the lowest error on the six patients that measured it, so it can name another pipeline.
With the privileged CT mask, its lowest pipeline ends 9.32~mm $[7.56, 11.08]$ above the pose oracle (6/6), not detectably below the CT-mask servo base ($-0.80$~mm $[-3.20, +1.60]$, 4/6; post hoc).
Selecting on five patients and scoring the sixth gives 11.87~mm, above the base's 11.65~mm.
Without privileged input, its post hoc lowest pipeline ends 9.91~mm $[7.39, 12.43]$ above the pose oracle (6/6) with label-conditioned frames.
No published component measured here, alone or in the rule's combination, closes the gap to the pose oracle.
The task thus remains under-explored, with much room for research.

\section{Controls, sensitivity analyses and replication}\label{sec:checks}

The analyses of this section ask whether the results of Sections~\ref{sec:results-ladder} and~\ref{sec:pluggability} depend on the testbed's settings or the six held-out patients (pre-specified criteria: Supplementary Table~S4.2; outcomes: Table~\ref{tab:checks-summary}); all but the primary real-frame results of Section~\ref{sec:results-real} are exploratory.

\begin{table*}[tp]
\renewcommand{\baselinestretch}{1}\scriptsize\setlength{\fboxsep}{0.6pt}
\caption{Controls, sensitivity analyses and replication: the question each answers and its outcome, absolute values first (exploratory). Supplementary Table~S4.2 gives each criterion. Every contrast is conditional on the trained artifacts (texture networks, segmenters, bases and policies). Seeds index artifacts; the patient or CT is the sampling unit. Intervals: 95\% paired $t$ with the patients sharing the sign. Criterion met: yes or no, inconclusive, descriptive, or, for a family of changes, how many differ.}
\label{tab:checks-summary}
\centering
\setlength{\tabcolsep}{3pt}
\begin{tabularx}{\textwidth}{@{}>{\raggedright\arraybackslash}p{0.13\textwidth} >{\raggedright\arraybackslash}p{0.18\textwidth} >{\raggedright\arraybackslash}X >{\raggedright\arraybackslash}p{0.085\textwidth}@{}}
\toprule
Analysis & Question & Outcome & Criterion met \\
\midrule
\multicolumn{4}{@{}l}{\textit{Controls}} \\
Independent target speckle & Are appearance scores insensitive to shared speckle? & Largest: learned-mask servo base, physics, seed~0, 16.39 against 17.03~mm, $-0.63$ $[-2.04, +0.78]$, 5/6; 4 conditions: $-0.63$ to $+0.05$~mm & yes \\[1pt]
Positive control, SB3 learner & Can the RL training loop learn a correction? & 2.40 to 2.58~mm in 3 of 3 seeds; bound 4.43, pose oracle 1.43, static start 15.45~mm & yes \\[1pt]
Positive control, REINFORCE & Can REINFORCE remove a planted deficit? & Planted lateral deficit 3.84~mm; 0.32 to 0.38 of it removed, positive in 0 of 6 patients for every setting; the planted bias did not meet the reward-visibility rule & inconclusive \\[1pt]
Label-channel controls & Does the SonoSPADE lead come from label access rather than texture? & Label-painted stage 14.22~mm against physics 17.20 and SonoSPADE 13.13~mm: $-2.97$ $[-5.50, -0.45]$, 6/6 from physics, 0.73 [0.33, 1.34] of the lead; equivalence with SonoSPADE not shown. Rendered masks follow offset and permuted labels, 6/6 each. Permuted organ identities 21.69~mm, $+8.56$ $[+4.46, +12.65]$, 6/6 & inconclusive \\[1pt]
Texture-budget test & Does the SonoSPADE lead persist with equal budgets and three texture seeds? & SonoSPADE, three texture seeds, 12.92~mm against budget-matched CUT 17.20, CycleGAN 16.93 and physics 17.20~mm: $-4.29$ $[-6.57, -2.00]$, 6/6; $-4.01$ $[-6.74, -1.27]$, 6/6; $-4.28$ $[-6.76, -1.80]$, 6/6~mm & yes \\[1pt]
Skin-tangent contact & Do the results depend on the target-anchored contact plane? & No level of the oracle ladder changes detectably ($+0.36$ to $+3.53$~mm, Holm $p$ $0.093$ to $0.620$); equivalence within $\pm 1$~mm is not shown. The CT-mask and learned-mask servo-base differences are $+3.53$ and $+3.00$~mm, 6/6 each. The plane lies near the target window (median angle 10.96$^\circ$). MDD 1.71~mm at the median SD; pose-oracle success at 5~mm and 5$^{\circ}$, $-62$ points (secondary) & 0 of 5 differ \\[1pt]
\addlinespace
\multicolumn{4}{@{}l}{\textit{Sensitivity}} \\
BatchNorm in eval mode & Are the results unchanged with eval-mode BatchNorm? & $-0.34$ to $+0.22$~mm over 18 policies; the outcomes of the 0.5~mm bound on imitation and REINFORCE and of the CT-mask stage-pair test change & no \\[1pt]
36-step horizon & Is the servo's error a horizon limit? & Servo base 18.34~mm at 36 steps against 11.65~mm at 12, $+6.69$ $[+4.96, +8.43]$, 6/6; pose oracle 0.14~mm & no \\[1pt]
Phantom bands & Do the renderers' properties lie in their designed ranges? & Bands met: physics renderer 5 of 8, original LOTUS code 2 of 4, SIMUS 5 of 8 (2, 2 and 1 not scorable). Outside: the disc CNR over the speckle SNR, 0.67 and 0.80 (band at least 0.90); the SIMUS shadow below the bone slab, 0.00~mm (band at least 10~mm). Output speckle SNR 3.57, 4.71 and 1.55, in that order (Rayleigh 1.91). Several bands are design checks & descriptive \\[1pt]
\addlinespace
\multicolumn{4}{@{}l}{\textit{Replication}} \\
20-CT cohort & Do the main contrasts recur beyond the six held-out patients? & Physics servo: CT mask 12.80, learned mask 15.36~mm; gap $+2.57$ $[+1.66, +3.47]$, 19/20, against $+4.92$, 5/6, on the six. SonoSPADE minus physics, learned-mask servo base: $-2.04$ $[-2.75, -1.33]$, 17/20, against $-4.06$, 5/6. RL change within 1~mm in all 8 configurations; the ordering recurs & yes \\[1pt]
Plugged methods on the cohort & Do plugged-method orderings recur on the cohort? & Orderings kept: controller 63 of 66, reward 25 of 28, renderer 3 of 3, mask source 16 of 21, texture 17 of 28; Kendall $\tau_b$ 0.21 [0.00, 0.64] to 1.00 [0.33, 1.00] & descriptive \\[1pt]
30 external CTs & Do the main contrasts recur on a second public CT dataset? & Learned minus CT masks, physics: $+4.01$ $[+3.40, +4.62]$, 29/30, against $+4.92$ on the six. SonoSPADE minus physics, learned-mask servo base: $-2.73$ $[-3.43, -2.03]$, 27/30, against $-4.06$. RL change within 1~mm in all 8 configurations; the ordering recurs. Simulation only & yes \\[1pt]
\bottomrule
\end{tabularx}
\end{table*}

\subsection{Training variability, inference mode and renderer phantoms}\label{sec:checks-repeat}

The 95\% prediction interval of the final error for one new training seed is $\pm 1.34$~mm.
With BatchNorm in eval mode, the CT-mask physics and per-tissue finals, the learned-mask CUT and CycleGAN rows after imitation and the learned-mask CUT final change by up to 0.34~mm.
Two pre-specified outcomes change: the 0.5~mm bound on imitation and REINFORCE in the learned-mask CUT and CycleGAN arms, and the CT-mask stage-pair test of physics against the per-tissue stage.
BatchNorm mode changes policy rows by more than the texture stages do under CT masks; both modes are reported (exploratory).

On canonical phantoms, the physics renderer, the original LOTUS code and SIMUS meet 5 of 8, 2 of 4 and 5 of 8 bands fixed before rendering, with 2, 2 and 1 not scorable (descriptive).
For the physics renderer and SIMUS, the anechoic disc's contrast-to-noise ratio over the speckle signal-to-noise ratio of phantom H, 0.67 and 0.80, falls below its minimum of 0.90.
SIMUS's shadow depth below the bone slab of phantom B, 0.00~mm, falls below its minimum of 10~mm.
Several bands are design checks that hold by the renderer's construction.
No renderer is called realistic.

\subsection{Positive controls of RL training}\label{sec:checks-controls}

\begin{figure*}[t]
\centering
\begin{minipage}[b]{0.325\textwidth}
\centering
\includegraphics[width=\textwidth]{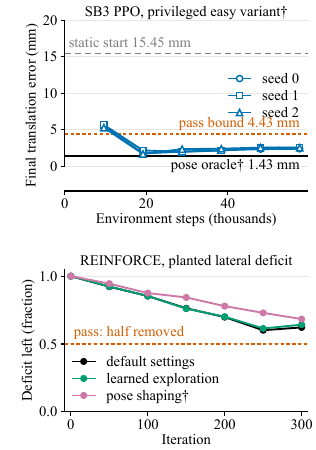}\\[2pt]
{\small (a)}
\end{minipage}\hfill
\begin{minipage}[b]{0.325\textwidth}
\centering
\includegraphics[width=\textwidth]{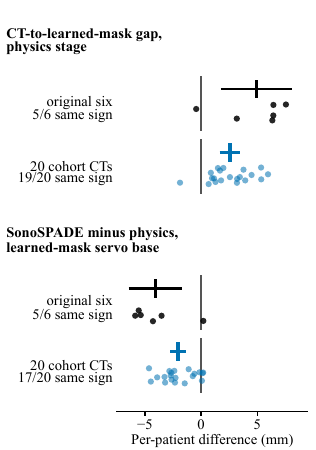}\\[2pt]
{\small (b)}
\end{minipage}\hfill
\begin{minipage}[b]{0.325\textwidth}
\centering
\includegraphics[width=\textwidth]{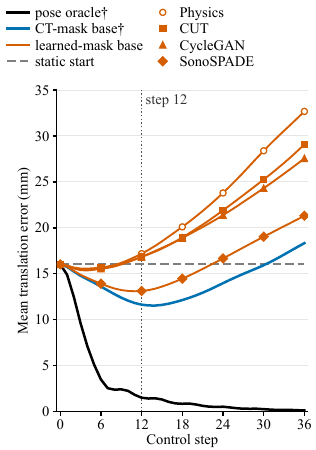}\\[2pt]
{\small (c)}
\end{minipage}
\caption{Controls and replication: exploratory results, except the original six patients in (b), which are primary; daggers mark privileged inputs. (a) Top: SB3 PPO on the privileged easy variant, final error on training-patient episodes against environment steps, three seeds, with the static start, pass bound and pose oracle. Bottom: the REINFORCE control, the fraction of the planted lateral deficit left against iteration, three settings, pass line at one half. (b) Per-patient differences of the two cohort contrasts, original six and 20 cohort CTs, with means and 95\% $t$ intervals, also for the original six, whose primary bootstrap intervals are in Table~\ref{tab:ladder}. (c) Mean translation error against control step 0 to 36; the dotted line marks the evaluation horizon, step 12. Every contrast is conditional on the trained artifacts (texture networks, segmenters, bases and policies). Seeds index artifacts; the patient is the sampling unit.}
\label{fig:checks}
\end{figure*}

On training patients, a PPO learner~\cite{sb3,schulman2017ppo} trained through the same environment and 57,600-step budget ends within 3~mm of the pose oracle on a privileged easy variant in all three seeds, so the RL training loop can learn a correction (exploratory).
REINFORCE~\cite{reinforce} does not recover half of a planted deficit under any of the three settings at 57,600 steps.
Nor does it recover half of a secondary deficit that halves the servo base's command at execution (recovery statistic positive in 1 of 6 patients for every setting).
The outcome is inconclusive: on training patients, the planted bias lowers the final match score by only 0.008, below the visibility rule's 0.03.
Raised to 1.0~mm per step, the largest value tried on TRAIN, the planted bias still lowers the final match score by only 0.023.
At this bias, REINFORCE removes 10\%, 9\% and 29\% of the 7.60~mm lateral deficit with default settings, learned exploration and privileged pose shaping (exploratory; estimation only; Supplementary S7).
Their patient-bootstrap intervals are [8\%, 12\%], [7\%, 11\%] and [27\%, 31\%], and the recovery statistic is positive in 0 of 6 patients for every setting.
The PPO control shows that the training loop can learn, not that learning from images is possible, because its learner, observation and reward differ from REINFORCE's (Fig.~\ref{fig:checks}a).
Every RL statement is scoped to this configuration and budget, and the RL null of Table~\ref{tab:ladder} says nothing about RL in general.

\subsection{Horizon and contact surface}\label{sec:checks-horizon}

At 36 steps, the CT-mask servo base ends at 18.34~mm and the learned-mask bases at 21.32 to 32.67~mm, above the 16.03~mm static start (Fig.~\ref{fig:checks}c).
Tripling the horizon changes the CT-mask servo error by $+6.69$~mm [$+4.96$, $+8.43$], 6/6; the 12-step horizon is not what holds the servo near 11.7~mm (exploratory).
At 36 steps the pose oracle ends at 0.14~mm; its 12-step residual of 1.53~mm is a horizon and gain effect (exploratory).
Post hoc, the logs trace this residual to lagged tracking: after each step, the oracle's probe trails its command by a mean 4.60~mm (Supplementary Table~S2.12).
Every plugged method is evaluated at 12 steps, which can hide later drift (Section~\ref{sec:plug-protocol}).

Replacing the target-anchored plane with a skin-tangent plane where the start pose's beam axis enters the skin, with no artifact retrained, changes no level of the oracle ladder detectably ($+0.36$ to $+3.53$~mm; Holm $p$ 0.093 to 0.620; MDD 1.71~mm at the experiment's median SD; Supplementary Table~\ref{tab:arm-b}).
The start is still drawn around the target pose, so this plane lies near the target window.
This experiment therefore measures a change of contact geometry, not contact free of target information.
Both servo bases keep the calibrations fit with the target-anchored plane.
Equivalence within $\pm 1$~mm is not shown for any level (TOST, Holm $p$ 0.337 to 1.000).
Both servo-base differences are positive in 6/6 patients, with unadjusted intervals above 0, [$+0.89$, $+6.18$] and [$+0.29$, $+5.72$]~mm.
In a secondary comparison, the pose oracle's success at 5~mm and 5$^{\circ}$ falls by 62 [46, 77] percentage points (Supplementary Table~S6.2).

\subsection{Label-channel controls and texture budget}\label{sec:checks-label}

Three controls ask whether the SonoSPADE lead comes from label access: out-of-plane label offsets, permuted organ identities and a label-painted stage.
These controls describe label-conditioned stages as a class; the published per-tissue method is used unchanged.
Permuting the organ identities raises the total final error from 13.13 to 21.69~mm, by $+8.56$~mm [$+4.46$, $+12.65$].
Label offsets of 3 and 6~mm change it by $+0.01$ and $+0.49$~mm.
Applied to the observations only, they shift the final elevational displacement by $-0.36$~mm [$-0.87$, $+0.14$] and $-0.82$~mm [$-1.52$, $-0.11$].
For both offsets, a shift of at least half the offset is not shown (exploratory).
With the label offset reversed ($-3$~mm), the shift is $+0.40$~mm [$-0.20$, $+1.01$] (secondary).
Given a label slice taken 3 or 6~mm out of plane, the per-tissue stage leads its matched segmenter to follow the given label rather than the anatomy in the physics input.
Its label-following index is $+0.142$ [$+0.116$, $+0.168$] and $+0.281$ [$+0.247$, $+0.315$], or 53\% and 64\% of complete following.
With organ identities permuted in the label, the masks follow the permuted identities ($+1.137$ [$+1.038$, $+1.235$], 6/6).
A non-learned label-painted physics stage lowers the learned-mask base error below physics ($-2.97$~mm; exploratory) and recovers 0.73 [0.33, 1.34] of the per-tissue lead.
Equivalence with the per-tissue stage is not shown ($n = 6$).
Its TOST within $\pm 1$~mm has a power of about 0.01.
The painted stage brightens the kidney from a depth-driven median of 1.89 toward a normal liver-to-kidney ratio of 1.14~\cite{shiralkar2015hri}, and draws 12~dB capsule lines, so its 0.73 mixes label access with renderer calibration.
Each label-control contrast that meets the decision rule has its sign in 6/6 patients and an exact sign-flip $p$ of 0.031 (paired $t$ and Holm $p$: 0.003 and 0.009 for the permuted total error, both at most $3.3 \times 10^{-5}$ for the masks, 0.029 for painting).

Neither the training budget nor a single texture seed explains the SonoSPADE lead.
We retrained CUT~\cite{cut} and CycleGAN~\cite{cyclegan}, which the reference pipeline trains on 1,540 frames for 800 iterations, at SonoSPADE's budget of 4,620 frames and 1,500 iterations, with an exponential moving average (EMA) of their weights.
With equal training budgets and three texture seeds per stage, the per-tissue stage gives a lower learned-mask base error than budget-matched CUT ($-4.29$~mm [$-6.57$, $-2.00$]), CycleGAN ($-4.01$ [$-6.74$, $-1.27$]) and physics ($-4.28$ [$-6.76$, $-1.80$]).
All three meet the decision rule, each in 6/6 patients (paired $t$ $p$ 0.005 to 0.013; Holm $p$ 0.014 and exact sign-flip $p$ 0.031 each).
Raising CUT and CycleGAN to the per-tissue budget changes their learned-mask base error by $+0.37$ and $+0.07$~mm (secondary).
The policy rows of Table~\ref{tab:ladder} still rest on one texture run per stage.

\subsection{Replication cohorts}\label{sec:checks-cohort}

The main contrasts of the six held-out patients recur on the 20 replication CTs, the mask and texture contrasts at about half their size.
With physics frames, the CT-mask and learned-mask servos end at 12.80 and 15.36~mm on these CTs.
These 20 cases come from the same public dataset (16 of 20 from institute E); this is internal replication, not external validity.
On 20 new CTs, learned masks with the physics stage add 2.57~mm over CT masks ([$+1.66$, $+3.47$], 19/20, Holm $p = 7.4 \times 10^{-5}$; exploratory), replicating the CT-to-learned-mask gap.
This gap is about half the $+4.92$~mm, 5/6, of the six held-out patients.
Both gaps are measured on the default render, whose organ boundaries fall below its display window.
The physics segmenters reach Dice 0.50 to 0.53 for liver and 0.17 to 0.42 for kidney on held-out simulated frames (post hoc; Supplementary Table~S2.9).
On real liver frames (Vitale holdout, AUL), they reach 0.58 to 0.69 and SonoSPADE's 0.40 to 0.58; on KidneyUS kidney, 0.35.
Each held-out simulated frame carries its own domain-randomization draw, so none is the default render.
On 20 new CTs, the per-tissue stage gives a lower learned-mask base error than physics ($-2.04$~mm [$-2.75$, $-1.33$], 17/20, Holm $p = 7.4 \times 10^{-5}$; exploratory).
This lead is about half the $-4.06$~mm, 5/6, of the six held-out patients (Fig.~\ref{fig:checks}b).
On 20 new CTs, REINFORCE fine-tuning changes the final error by less than 1~mm in every policy arm (TOST, intersection-union, Holm $p = 7.4 \times 10^{-5}$; exploratory).
Applied post hoc to the six held-out patients, a TOST at the same margin shows equivalence in 8 of 8 policy configurations, but the crossed patient-by-seed interval with CUT frames and learned masks reaches $+1.18$~mm (Supplementary Table~S2.7).
The intersection-union form of the cohort test was fixed when this post hoc result on the six held-out patients was already known.
The rung and stage ordering of the original six recurs on 20 new CTs (100\% of the 40 rule-holding pairs keep their sign; Kendall $\tau_b = 0.61$ [0.45, 0.82]).
On 20 new CTs, the servo base under CT masks ends 11.17~mm farther from the target than the privileged pose oracle (oracle-to-CT-mask gap; exploratory).
On 20 new CTs, privileged look-ahead ascent on the CT-mask match score (a privileged reference rung) ends below the servo base by 4.96~mm [3.68, 6.23] (exploratory).
Both meet the cohort decision rule, with the same sign in 20/20 CTs (exact sign-flip $p = 1.9 \times 10^{-6}$; Holm $p = 3.8 \times 10^{-6}$).
Both are conditional on this renderer: grid-fixed speckle, no elevational beam width, rigid tissue.
This bound uses look-ahead sampling seed 0, as pre-specified.
With sampling seeds 0 to 2 averaged, it ends at 7.51~mm, 5.13~mm [3.87, 6.39] below the servo base (20/20; exploratory; estimation only; conditional on this renderer; Supplementary S7).

A second cohort of 30 CTs from the public AMOS 2022 dataset~\cite{ji2022amos}, with TotalSegmentator v2 labels checked against its manual organ labels and every trained artifact unchanged, replicates the same contrasts (\ref{app:decisions-amos}).
Learned masks with the physics stage add 4.01~mm over CT masks (29/30), SonoSPADE frames give the learned-mask servo base a lower error than physics frames ($-2.73$~mm, 27/30), REINFORCE fine-tuning changes the final error by less than 1~mm in every policy configuration, and the ordering of rungs and stages recurs.
This is external replication in simulation only: CT source and label source change together, and no AMOS result speaks to real ultrasound.
Whether TotalSegmentator v2 was trained on AMOS images is not recorded, so the labels are treated as in-sample predictions; no artifact of the testbed was trained on an AMOS CT.
On 30 AMOS CTs, the servo base under CT masks ends 10.68~mm farther from the target than the privileged pose oracle (oracle-to-CT-mask gap; exploratory).
On 30 AMOS CTs, privileged look-ahead ascent on the CT-mask match score (a privileged reference rung) ends below the servo base by 4.51~mm [3.74, 5.29] (exploratory).
Both hold in 30/30 CTs, with Monte Carlo sign-flip $p = 1.0 \times 10^{-6}$ (1,000,000 sign vectors, seed 0) and Holm $p = 2.0 \times 10^{-6}$.
Both are conditional on this renderer, as on the 20 CTs.
This bound uses look-ahead sampling seed 0, as pre-specified.
With sampling seeds 0 to 2 averaged, it ends at 7.40~mm, 4.77~mm [4.05, 5.49] below the servo base (29/30; exploratory; estimation only; conditional on this renderer; Supplementary S7).
\ref{app:decisions-amos} reports, as estimation only, a relabeling control on the 20 replication CTs, a comparison of the two cohorts and the secondary results.

\subsection{Offline evaluation on real frames}\label{sec:results-real}

On public real frames, a classifier distinguishes every texture stage from real images, and selection by the match score gives small organ-overlap gains for some combinations of stage, dataset and organ; this offline check uses no probe, phantom or robot.
The check does not show directed control on real frames (Table~\ref{tab:real-cells}).
Against 60 annotated public frames~\cite{vitale} that no stage trained on, the two-sample classifier's AUC~\cite{c2st} is 0.701 for CUT, 0.753 for CycleGAN, 0.936 for physics and 0.979 for SonoSPADE (Supplementary Table~\ref{tab:realism}).
In a post hoc label-permutation null with 200 permutations per stage, the AUC averages 0.50 (SD 0.07) and never reaches the observed value of any stage ($p = 1/201$ each; Supplementary Table~S2.16).
The physics renderer is closer to real than the authors' reimplementation of LOTUS, with a pooled W1 of 0.0727 against 0.1272 and a lower W1 in 6/6 patients.
In a post hoc comparison, the physics renderer's W1 is also below that of the original LOTUS code, 0.0875, pooled and in 5/6 patients.
This W1 is sensitive to display settings, because one domain-randomization draw per view lowers it to 0.0154.
The 0.77 sector compression of Section~\ref{sec:rendering} enters every real-frame metric and the matched-layout ceiling, the median IoU, 0.56, of 60 real frames against their best-matching held-out CT views.
No stage is called realistic.
Organ identity counts the scored organs whose most frequent predicted class is correct in both domains.
It shows no difference detected across stages on all 60 Vitale frames and on the 18 holdout frames (not powered for equivalence; Supplementary Table~\ref{tab:identity}).
\begin{table*}[tbp]
\centering
\renewcommand{\baselinestretch}{1}\scriptsize
\setlength{\tabcolsep}{3pt}
\caption{Outcomes of the real-frame cells of SonoSPADE, the standard engine's texture stage (all eight testable cells, with the ladder's other texture settings: Supplementary Table~\ref{tab:real-cells-all}). Keep-best gain: primary gain over the static start (Supplementary Table~\ref{tab:registration}). Selection: organ IoU of the state the match score selects along undirected motion, minus that of a random state. Directed control: the policy's keep-best gain minus that of magnitude-matched undirected motion with the same selection. Values: exploratory per-subject IoU means [95\% subject-by-CT bootstrap CI]. A cell holds if its CI excludes 0 and its Holm $p$ over eight cells is below 0.05. In the holding cell, 38/55 subjects share the sign; exact sign-flip $p = 3.4 \times 10^{-4}$ and Holm $p = 0.003$. Elsewhere Holm $p$ is 0.055 to 1.000.}
\label{tab:real-cells}
\begin{tabularx}{\textwidth}{@{}l>{\raggedright\arraybackslash\hsize=0.95\hsize}Xl>{\raggedright\arraybackslash\hsize=1.0\hsize}Xl>{\raggedright\arraybackslash\hsize=1.05\hsize}X@{}}
\toprule
Stage & Keep-best gain & Selection [CI] & Outcome & Directed control [CI] & Outcome \\
\midrule
\multicolumn{6}{@{}l}{\textit{Vitale holdout~\cite{vitale}, liver, 12 subjects}} \\
\quad SonoSPADE & moves the view & $+0.011$ [$+0.002$, $+0.024$] & not detectably better & $+0.041$ [$-0.001$, $+0.098$] & selection, not directed control \\
\multicolumn{6}{@{}l}{\textit{KidneyUS~\cite{kidneyus}, kidney, 55 subjects}} \\
\quad SonoSPADE & moves the view & $+0.015$ [$+0.003$, $+0.028$] & informative selection signal & $+0.014$ [$-0.011$, $+0.037$] & selection, not directed control \\
\bottomrule
\end{tabularx}
\end{table*}

The reward, computed on real frames, moves the simulated view toward the real target's organ layout for liver, and for kidney with SonoSPADE on KidneyUS.
Elsewhere there is no reliable improvement on real targets, including every stage on the 4 Vitale kidney targets.
Keep-best gains over the static start have CIs above 0 in 7 of 16 combinations (Supplementary Table~\ref{tab:registration}), and physics on KidneyUS kidney falls, at $-0.014$ [$-0.026$, $-0.001$].
Under a post hoc bootstrap crossed over targets, CTs and seeds, only SonoSPADE on Vitale-holdout liver keeps a CI above 0, and none also meets a Holm-adjusted $t$-test (Supplementary S2).
Absolute keep-best IoU is 0.303 to 0.344 on Vitale-holdout liver and 0.144 to 0.186 on KidneyUS kidney~\cite{kidneyus}, against a matched-layout ceiling of 0.56 (exploratory).
Final-step gains are $-0.073$ to $-0.031$ on KidneyUS kidney, negative for all four stages, and $+0.016$ to $+0.050$ on Vitale-holdout liver (exploratory).
Real-side units are frames or file-stem groups of unknown patient identity and may not be independent.

\section{Discussion}\label{sec:discussion}

\subsection{Principal findings}\label{sec:taught}
The answer to the research question is no: in this testbed, published components plugged one slot at a time did not close the gap to the pose oracle.
Without privileged input, no level of the ladder succeeded at 5~mm and 5$^{\circ}$ in over 5.6\% of episodes (post hoc).
Among controllers, only the DAgger residual separated from its base ($-3.17$~mm, 6/6; Section~\ref{sec:plug-ctrl}), a base not detectably closer than the start.
In the primary analysis, most of the error ($+10.1$ to $+10.3$~mm, 6/6) lies between the pose oracle and the CT-mask servo.
The oracle-to-CT-mask gap is not attributed to one cause (conditional on this renderer: grid-fixed speckle, no elevational beam width, rigid tissue).
The CT-to-learned-mask gap is the second limit, at $+4.80$ to $+5.26$~mm for three of four stages on the nearly isoechoic default render.
Across the engine's texture settings, the smaller CT-to-learned-mask gap went with label access, not with distance to real frames by KID.
As the label-painted control of Section~\ref{sec:checks-label} shows, label access is a property of the engine's texture stage, not of one method.

\subsection{Using the testbed and reporting new methods}\label{sec:use-report}
Unlike hardware tests~\cite{hu2026tmrb,qian2026}, USB Gym compares components slot by slot against privileged references, in simulation only.
A new method can be reported as in Table~\ref{tab:pluggability}, from at least three training seeds with the patient-by-seed interval of \ref{app:decisions}.
One new reference seed is expected within $\pm 1.34$~mm of the seed mean (95\% prediction interval; Section~\ref{sec:checks-repeat}).
At $n = 6$, a plugged method's result is an estimate with its interval, read beside coverage and cost, not a ranking.
Of 19 contrasts in the guide's slots, 15 showed no detectable difference; the task remains under-explored.

\subsection{Limitations}\label{sec:limitations}
USB Gym cannot show how tissue deforms under pressure or whether a rendered frame looks real.
Deformable ultrasound simulation~\cite{goksel2009,burger2013,starkov2019} offers a route to pressure-dependent images.
By default, the renderer has grid-fixed speckle, no elevational beam width and a nearly isoechoic default.
The reference renderer was compared with no physical phantom, only with the original LOTUS code, SIMUS and digital phantom bands.
The real-frame results score the reward offline, so we claim no sim-to-real transfer of a policy.

Every absolute error and success rate is conditional on the target-anchored contact plane, one per target window and normal to its beam.
The skin-tangent contact plane lies near the target window, so contact still carries target information (Section~\ref{sec:checks-horizon}); deformable contact with the CT skin itself could remove this dependence.
Under the servo base and pose oracle, the probe lags its command by 1.3 to 1.5 mean commanded steps (post hoc; Supplementary Table~S2.12).
Neither the contact nor the actuation model has been validated against a robot.
The loop models no respiration, arm kinematics, tracking noise, contact force in the observation or loss of gel coupling.

Without tuning, no result ranks a method beyond this configuration or for clinical use.
The reward weights, the action cost and the DR ranges were set by hand and never varied, and the DR landscape has intervals over three segmenter seeds only (post hoc).
The learned-mask controller experiment removes privileged input from the controller and mask source only (Section~\ref{sec:plug-ctrl}).
The v2 relabeling (\ref{sec:plug-anatomy}) uses TotalSegmentator's own training CTs, so it likely understates an independent segmenter's label noise.
In the AMOS CTs, CT source and label source change together (Section~\ref{sec:checks-cohort}).
Every result was computed on one Apple-silicon laptop.
Evaluating the learned-mask servo base on the CPU instead of the laptop's GPU changes a patient's final error by up to 0.22~mm (physics stage, seed 0; exploratory), more than the smallest reported differences, so small contrasts should be compared on one device.

SonoSPADE~\cite{sonospade} was developed by three of the authors and was used unchanged, so the comparisons against it are not independent of its authors.
The guide's slots were compared on the standard engine only, whose SonoSPADE frames carry label information (Section~\ref{sec:checks-label}).

\section{Conclusions}\label{sec:conclusion}
Plugged one slot at a time, untuned and at the reference training budget, existing published components did not close the gap to the pose oracle in this simulator.
The guide was compared on the standard simulation engine, a physics renderer with the SonoSPADE texture stage, which stays pluggable.
To test the testbed, the combination of slot components chosen on six held-out patients, our pipeline with an SSIM reward term, was tested on 20 new CTs.
On the standard engine, it was the benchmark baseline, at 13.70~mm (pose oracle 1.47~mm; exploratory; conditional on this renderer).
Its 0.26~mm lead over our reference pipeline lies within $\pm 1$~mm and is not attributed to SSIM (Section~\ref{sec:plug-best}).
Of 19 contrasts in the guide's slots, 4 differed detectably and 15 did not, 2 of these within $\pm 1$~mm (post hoc) and 13 inconclusive at $n = 6$ (exploratory).
The task therefore remains under-explored, with much room for research.
The testbed accepts new components in each slot.
No policy ran on hardware, so a phantom or robot test is the next step.

\section*{Data and Software Availability}
The code, tests and documentation are public at \url{https://github.com/tohnperfect/USBGym} under the Apache License 2.0.
Zenodo archives every release~\cite{usbgym} under the concept DOI \href{https://doi.org/10.5281/zenodo.23942190}{10.5281/zenodo.23942190}, which resolves to the latest; that release's \texttt{scripts/paper} folder rebuilds every table and figure data file of the article and the supplement from the result files below and compares each with the printed one, except the figure data that need the texture-network checkpoints or the Vitale frames, which the result files leave out.
Every printed number can be recomputed from it, and retraining reproduces a result only within the training variability of Section~\ref{sec:checks-repeat}.
The segmenter training code is in the \texttt{segtrain} folder of the code repository.
A second Zenodo record~\cite{usbgymdata} holds the result files behind every table, under DOI \href{https://doi.org/10.5281/zenodo.23913142}{10.5281/zenodo.23913142}.
\ref{app:leaderboard} lists every plugged method of each slot from these files (Supplementary Table~\ref{tab:leaderboard}).

All data behind the results come from public datasets.
CT anatomy, including the 20 replication CTs, comes from the TotalSegmentator dataset~\cite{totalseg} under CC BY 4.0.
The 30 CTs of the external cohort come from AMOS 2022~\cite{ji2022amos}.
Its dataset description states CC BY-SA 4.0 and its Zenodo record CC BY 4.0, so any derivative is released, if at all, under CC BY-SA 4.0 with attribution.
Real abdominal frames come from the public Kaggle set of Vitale et al.~\cite{vitale}, whose page reserves the data files to their original authors and asks users to cite~\cite{vitale}.
Real kidney frames come from KidneyUS~\cite{kidneyus} under \mbox{CC BY-NC-SA}.
Its maintainers grant access only after registration and manual approval.
Segmenter weights trained on KidneyUS are not released; the \texttt{segtrain} folder of the code repository retrains them from the datasets above.
Real liver frames come from the Annotated Ultrasound Liver dataset~\cite{aul,auldata} under CC BY 4.0.
The liver frames, the KidneyUS holdout split and a group-disjoint Vitale holdout serve only as real targets.
\newif\ifvitaleframes
\ifdefined\novitale\vitaleframesfalse\else\vitaleframestrue\fi
\ifvitaleframes
Supplementary Fig.~\ref{fig:layout} reproduces frames of the public Vitale set~\cite{vitaledata}.
\else
Supplementary Fig.~\ref{fig:layout} shows the matched CT views of frames of the public Vitale set~\cite{vitaledata} without reproducing those frames.
\fi

The plugged methods use third-party code and weights under their own licenses.
Supplementary S3 lists each license.

\section*{Ethics Statement}
The results use only public, de-identified datasets.
No new human or animal data were collected.
No experiment involved human participants.
Ethical approval was therefore not sought.

\section*{Author Contributions}
Thanapong Intharah contributed conceptualization, methodology, software, validation, formal analysis, investigation, data curation, visualization, writing (original draft), writing (review and editing), project administration and funding acquisition.
Zhichao Gao contributed conceptualization and methodology.
Yiliang Liu contributed conceptualization and validation.
Hao Dong contributed supervision.

\section*{Conflict of Interest}
Three of the authors, T.~Intharah, Z.~Gao and H.~Dong, developed SonoSPADE~\cite{sonospade}.
This study uses SonoSPADE, with its method unchanged, as the texture stage of its standard simulation engine.
The authors declare no other known competing financial interests or personal relationships that could have appeared to influence the work reported in this paper.

\section*{Acknowledgment}
During the preparation of this work the authors used Claude Code (Anthropic)~\cite{claudecode} in order to draft and edit the text, write and test the testbed and analysis code, and run the experiments and analyses.
The tool ran the models Claude Opus 4.8, Claude Opus 5 and Claude Opus 5.5, as the commit history of the code records.
It also checked the references against Crossref and PubMed. The authors reviewed the text in successive drafts and checked the code through its test suite, run on Linux, Windows and macOS, and through the scripts that rebuild every table and figure data file from the archived results.
After using this tool, the authors reviewed and edited the content as needed and take full responsibility for the content of the published article.

\begin{IEEEbiography}[{\includegraphics[width=1in,height=1.25in,clip,keepaspectratio]{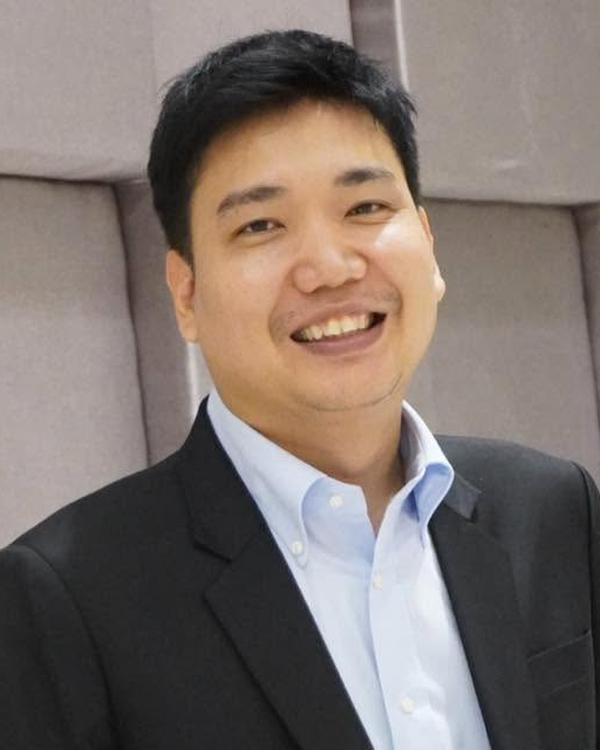}}]{THANAPONG INTHARAH}
(Member, IEEE) received the B.Sc. degree in mathematics from Mahidol University, Thailand, the M.Sc. degree in computer science from Chulalongkorn University, Thailand, the M.Sc. degree in machine learning from University College London, U.K., and the Ph.D. degree in computer science from University College London, in 2018.

He is currently an Associate Professor in computer science with the Department of Statistics, Faculty of Science, Khon Kaen University, Thailand. His research interests include applications of AI, machine learning, and computer vision in science and medicine.
\end{IEEEbiography}

\begin{IEEEbiography}[{\includegraphics[width=1in,height=1.25in,clip,keepaspectratio]{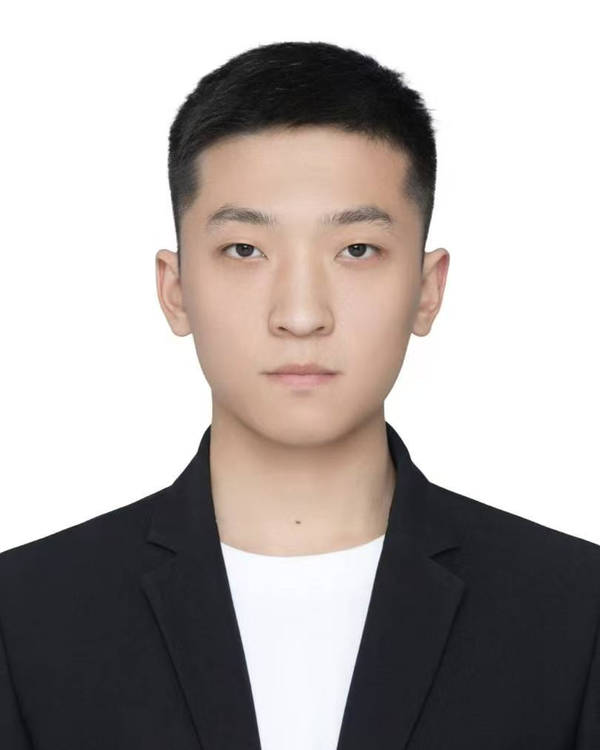}}]{ZHICHAO GAO}
received the bachelor's degree from Qingdao University of Science and Technology, Qingdao, China, and the master's degree from Beijing University of Posts and Telecommunications, Beijing, China.

He is currently a Research Assistant with the School of Computer Science, Peking University, Beijing, China, working under the supervision of Prof. Hao Dong. His research interests include robotics and embodied intelligence.
\end{IEEEbiography}

\begin{IEEEbiography}[{\includegraphics[width=1in,height=1.25in,clip,keepaspectratio]{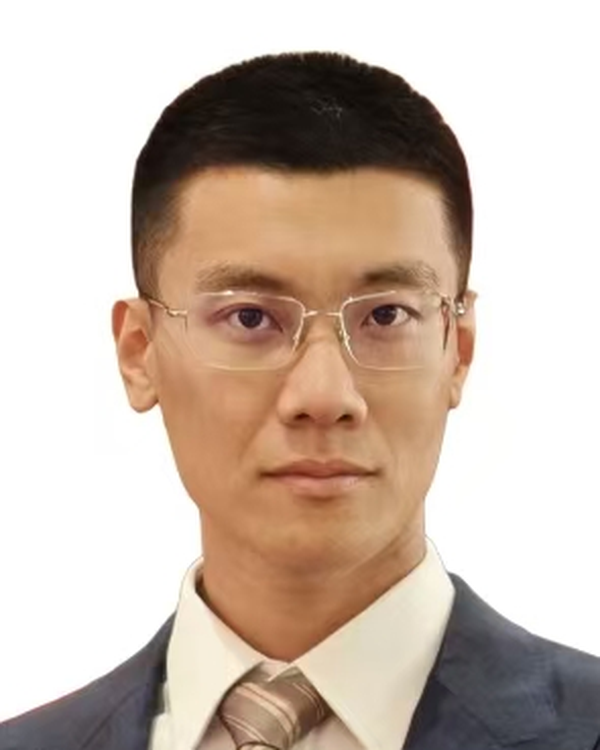}}]{YILIANG LIU}
received the B.A. degree in Arabic from Beijing International Studies University, Beijing, China.

He is currently an Artificial Intelligence Engineer with the School of Computer Science, Peking University, Beijing, China, working under the supervision of Prof. Hao Dong. His research interests include physical AI and autonomous agents. In addition, he is responsible for industrial collaboration and international exchange within the group.
\end{IEEEbiography}

\begin{IEEEbiography}[{\includegraphics[width=1in,height=1.25in,clip,keepaspectratio]{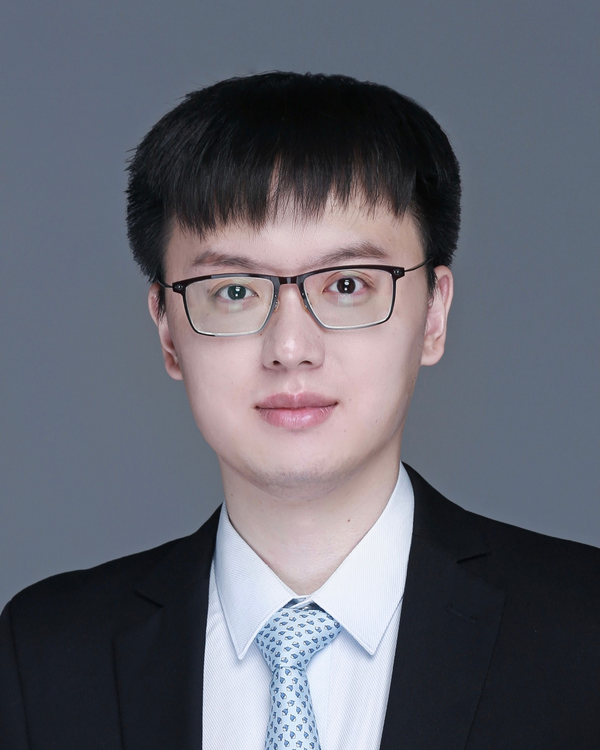}}]{HAO DONG}
received the B.S. degree from the University of Lancashire, U.K., and the M.Sc. and Ph.D. degrees with distinction from Imperial College London, U.K.

He is currently a tenured Associate Professor with the School of Computer Science and the Center for Frontiers in Computing Science, Peking University, Beijing, China, and the Chief Scientist of QYRO. He is a laureate of the National Young Talent Program. His research explores scaling laws for embodied intelligence, with the goal of building cost-effective, general-purpose and versatile robotic systems.
\end{IEEEbiography}

\EOD


\begin{thebibliography}{99}

\bibitem{vonhaxthausen2021}
F.~von~Haxthausen, S.~B\"{o}ttger, D.~Wulff, J.~Hagenah, V.~Garc\'{\i}a-V\'{a}zquez, and S.~Ipsen, ``Medical robotics for ultrasound imaging: current systems and future trends,'' \emph{Curr. Robot. Rep.}, vol.~2, pp.~55--71, 2021, doi: \href{https://doi.org/10.1007/s43154-020-00037-y}{10.1007/s43154-020-00037-y}.

\bibitem{gymnasium}
M.~Towers \emph{et~al.}, ``Gymnasium: a standard interface for reinforcement learning environments,'' in \emph{Advances in Neural Information Processing Systems 38 (NeurIPS 2025)}, 2025, pp.~163114--163129, doi: \href{https://doi.org/10.52202/085713-4916}{10.52202/085713-4916}.

\bibitem{totalseg}
J.~Wasserthal \emph{et~al.}, ``TotalSegmentator: robust segmentation of 104 anatomic structures in CT images,'' \emph{Radiol. Artif. Intell.}, vol.~5, Art.~no.~e230024, 2023, doi: \href{https://doi.org/10.1148/ryai.230024}{10.1148/ryai.230024}.

\bibitem{sonospade}
T.~Intharah, Z.~Gao, and H.~Dong, ``SonoSPADE: real-time, per-tissue ultrasound texture synthesis for closed-loop acquisition,'' in \emph{Medical Image Computing and Computer Assisted Intervention, MICCAI 2026 Workshops and Challenges (SASHIMI)}, ser. Lect. Notes Comput. Sci., vol.~17258, 2026. [Online]. Available: \url{https://papers.miccai.org/miccai-2026-sat/SASHIMI_035.html}

\bibitem{usbgym}
T.~Intharah, Z.~Gao, Y.~Liu, and H.~Dong, ``USB Gym: a Gymnasium testbed for closed-loop ultrasound view matching (software, all versions),'' Zenodo, 2026, license Apache 2.0, doi: \href{https://doi.org/10.5281/zenodo.23942190}{10.5281/zenodo.23942190}. Code repository: \url{https://github.com/tohnperfect/USBGym}.

\bibitem{kutter2009}
O.~Kutter, R.~Shams, and N.~Navab, ``Visualization and GPU-accelerated simulation of medical ultrasound from CT images,'' \emph{Comput. Methods Programs Biomed.}, vol.~94, pp.~250--266, 2009, doi: \href{https://doi.org/10.1016/j.cmpb.2008.12.011}{10.1016/j.cmpb.2008.12.011}.

\bibitem{salehi2015}
M.~Salehi, S.-A.~Ahmadi, R.~Prevost, N.~Navab, and W.~Wein, ``Patient-specific 3D ultrasound simulation based on convolutional ray-tracing and appearance optimization,'' in \emph{Medical Image Computing and Computer-Assisted Intervention, MICCAI 2015}, ser. Lect. Notes Comput. Sci., vol.~9350, 2015, pp.~510--518, doi: \href{https://doi.org/10.1007/978-3-319-24571-3_61}{10.1007/978-3-319-24571-3\_61}.

\bibitem{jensen1992field}
J.A.~Jensen and N.B.~Svendsen, ``Calculation of pressure fields from arbitrarily shaped, apodized, and excited ultrasound transducers,'' \emph{IEEE Trans. Ultrason. Ferroelectr. Freq. Control}, vol.~39, pp.~262--267, 1992, doi: \href{https://doi.org/10.1109/58.139123}{10.1109/58.139123}.

\bibitem{treeby2010kwave}
B.E.~Treeby and B.T.~Cox, ``k-Wave: MATLAB toolbox for the simulation and reconstruction of photoacoustic wave fields,'' \emph{J. Biomed. Opt.}, vol.~15, Art.~no.~021314, 2010, doi: \href{https://doi.org/10.1117/1.3360308}{10.1117/1.3360308}.

\bibitem{garcia2022simus}
D.~Garcia, ``SIMUS: an open-source simulator for medical ultrasound imaging. Part I: theory \& examples,'' \emph{Comput. Methods Programs Biomed.}, vol.~218, Art.~no.~106726, 2022, doi: \href{https://doi.org/10.1016/j.cmpb.2022.106726}{10.1016/j.cmpb.2022.106726}.

\bibitem{cigier2022simus}
A.~Cigier, F.~Varray, and D.~Garcia, ``SIMUS: an open-source simulator for medical ultrasound imaging. Part II: comparison with four simulators,'' \emph{Comput. Methods Programs Biomed.}, vol.~220, Art.~no.~106774, 2022, doi: \href{https://doi.org/10.1016/j.cmpb.2022.106774}{10.1016/j.cmpb.2022.106774}.

\bibitem{velikova2023lotus}
Y.~Velikova, M.F.~Azampour, W.~Simson, V.~Gonzalez~Duque, and N.~Navab, ``LOTUS: learning to optimize task-based US representations,'' in \emph{Medical Image Computing and Computer Assisted Intervention, MICCAI 2023}, ser. Lect. Notes Comput. Sci., vol.~14220, 2023, pp.~435--445, doi: \href{https://doi.org/10.1007/978-3-031-43907-0_42}{10.1007/978-3-031-43907-0\_42}.

\bibitem{cactuss}
Y.~Velikova, W.~Simson, M.~Salehi, M.F.~Azampour, P.~Paprottka, and N.~Navab, ``CACTUSS: common anatomical CT-US space for US examinations,'' in \emph{Medical Image Computing and Computer Assisted Intervention, MICCAI 2022}, ser. Lect. Notes Comput. Sci., vol.~13433, 2022, pp.~492--501, doi: \href{https://doi.org/10.1007/978-3-031-16437-8_47}{10.1007/978-3-031-16437-8\_47}.

\bibitem{cyclegan}
J.-Y.~Zhu, T.~Park, P.~Isola, and A.A.~Efros, ``Unpaired image-to-image translation using cycle-consistent adversarial networks,'' in \emph{Proc. IEEE Int. Conf. Comput. Vis. (ICCV)}, 2017, pp.~2242--2251, doi: \href{https://doi.org/10.1109/ICCV.2017.244}{10.1109/ICCV.2017.244}.

\bibitem{vitale}
S.~Vitale, J.I.~Orlando, E.~Iarussi, and I.~Larrabide, ``Improving realism in patient-specific abdominal ultrasound simulation using CycleGANs,'' \emph{Int. J. Comput. Assist. Radiol. Surg.}, vol.~15, pp.~183--192, 2020, doi: \href{https://doi.org/10.1007/s11548-019-02046-5}{10.1007/s11548-019-02046-5}.

\bibitem{spade}
T.~Park, M.-Y.~Liu, T.-C.~Wang, and J.-Y.~Zhu, ``Semantic image synthesis with spatially-adaptive normalization,'' in \emph{Proc. IEEE/CVF Conf. Comput. Vis. Pattern Recognit. (CVPR)}, 2019, pp.~2332--2341, doi: \href{https://doi.org/10.1109/CVPR.2019.00244}{10.1109/CVPR.2019.00244}.

\bibitem{jiang2023}
Z.~Jiang, S.E.~Salcudean, and N.~Navab, ``Robotic ultrasound imaging: state-of-the-art and future perspectives,'' \emph{Med. Image Anal.}, vol.~89, Art.~no.~102878, 2023, doi: \href{https://doi.org/10.1016/j.media.2023.102878}{10.1016/j.media.2023.102878}.

\bibitem{droste2020}
R.~Droste, L.~Drukker, A.T.~Papageorghiou, and J.A.~Noble, ``Automatic probe movement guidance for freehand obstetric ultrasound,'' in \emph{Medical Image Computing and Computer Assisted Intervention, MICCAI 2020}, ser. Lect. Notes Comput. Sci., vol.~12263, 2020, pp.~583--592, doi: \href{https://doi.org/10.1007/978-3-030-59716-0_56}{10.1007/978-3-030-59716-0\_56}.

\bibitem{li2021}
K.~Li \emph{et~al.}, ``Autonomous navigation of an ultrasound probe towards standard scan planes with deep reinforcement learning,'' in \emph{Proc. IEEE Int. Conf. Robot. Autom. (ICRA)}, 2021, pp.~8302--8308, doi: \href{https://doi.org/10.1109/ICRA48506.2021.9561295}{10.1109/ICRA48506.2021.9561295}.

\bibitem{amadou2024}
A.A.~Amadou \emph{et~al.}, ``Goal-conditioned reinforcement learning for ultrasound navigation guidance,'' in \emph{Medical Image Computing and Computer Assisted Intervention, MICCAI 2024}, ser. Lect. Notes Comput. Sci., vol.~15011, 2024, pp.~319--329, doi: \href{https://doi.org/10.1007/978-3-031-72120-5_30}{10.1007/978-3-031-72120-5\_30}.

\bibitem{hu2024spie}
Y.~Hu \emph{et~al.}, ``Probe positioning for robot-assisted intraoperative ultrasound imaging using deep reinforcement learning,'' in \emph{Medical Imaging 2024: Image-Guided Procedures, Robotic Interventions, and Modeling}, ser. Proc. SPIE, vol.~12928, 2024, Art.~no.~1292803, doi: \href{https://doi.org/10.1117/12.3006918}{10.1117/12.3006918}.

\bibitem{hu2026tmrb}
Y.~Hu \emph{et~al.}, ``Autonomous localization of renal ultrasound scan planes using sim-to-real reinforcement learning with image-based rewards,'' \emph{IEEE Trans. Med. Robot. Bionics}, vol.~8, pp.~1005--1014, 2026, doi: \href{https://doi.org/10.1109/TMRB.2026.3699227}{10.1109/TMRB.2026.3699227}.

\bibitem{han2026}
T.~Han \emph{et~al.}, ``Automatic probe guidance for volumetric liver ultrasound acquisition via imitation learning from a virtual expert,'' \emph{Int. J. Comput. Assist. Radiol. Surg.}, 2026, published online 6~September 2026, doi: \href{https://doi.org/10.1007/s11548-026-03787-w}{10.1007/s11548-026-03787-w}.

\bibitem{qian2026}
Y.~Qian \emph{et~al.}, ``Task-relevant feature-dynamics fidelity enables zero-shot sim-to-real transfer for robotic ultrasound scanning,'' 2026, arXiv:2608.29516.

\bibitem{mebarki2010}
R.~Mebarki, A.~Krupa, and F.~Chaumette, ``2-D ultrasound probe complete guidance by visual servoing using image moments,'' \emph{IEEE Trans. Robot.}, vol.~26, pp.~296--306, 2010, doi: \href{https://doi.org/10.1109/TRO.2010.2042533}{10.1109/TRO.2010.2042533}.

\bibitem{krupa2009}
A.~Krupa, G.~Fichtinger, and G.D.~Hager, ``Real-time motion stabilization with B-mode ultrasound using image speckle information and visual servoing,'' \emph{Int. J. Robot. Res.}, vol.~28, pp.~1334--1354, 2009, doi: \href{https://doi.org/10.1177/0278364909104066}{10.1177/0278364909104066}.

\bibitem{hase2020}
H.~Hase \emph{et~al.}, ``Ultrasound-guided robotic navigation with deep reinforcement learning,'' in \emph{Proc. IEEE/RSJ Int. Conf. Intell. Robots Syst. (IROS)}, 2020, pp.~5534--5541, doi: \href{https://doi.org/10.1109/IROS45743.2020.9340913}{10.1109/IROS45743.2020.9340913}.

\bibitem{bi2022vesnet}
Y.~Bi, Z.~Jiang, Y.~Gao, T.~Wendler, A.~Karlas, and N.~Navab, ``VesNet-RL: simulation-based reinforcement learning for real-world US probe navigation,'' \emph{IEEE Robot. Autom. Lett.}, vol.~7, pp.~6638--6645, 2022, doi: \href{https://doi.org/10.1109/LRA.2022.3176112}{10.1109/LRA.2022.3176112}.

\bibitem{sonogym}
Y.~Ao \emph{et~al.}, ``SonoGym: high performance simulation for challenging surgical tasks with robotic ultrasound,'' in \emph{Advances in Neural Information Processing Systems 38 (NeurIPS 2025)}, 2025, pp.~45013--45040, doi: \href{https://doi.org/10.52202/085713-1343}{10.52202/085713-1343}.

\bibitem{isaaclab}
M.~Mittal \emph{et~al.}, ``Orbit: a unified simulation framework for interactive robot learning environments,'' \emph{IEEE Robot. Autom. Lett.}, vol.~8, pp.~3740--3747, 2023, doi: \href{https://doi.org/10.1109/LRA.2023.3270034}{10.1109/LRA.2023.3270034}. Isaac Lab development started from the Orbit framework.

\bibitem{pix2pix}
P.~Isola, J.-Y.~Zhu, T.~Zhou, and A.A.~Efros, ``Image-to-image translation with conditional adversarial networks,'' in \emph{Proc. IEEE Conf. Comput. Vis. Pattern Recognit. (CVPR)}, 2017, pp.~5967--5976, doi: \href{https://doi.org/10.1109/CVPR.2017.632}{10.1109/CVPR.2017.632}.

\bibitem{i4h}
NVIDIA Corporation, ``Isaac for Healthcare: i4h-workflows and i4h-sensor-simulation,'' GitHub repositories, license Apache-2.0, 2026. [Online]. Available: \url{https://github.com/isaac-for-healthcare/i4h-workflows}; \url{https://github.com/isaac-for-healthcare/i4h-sensor-simulation}. Accessed: 19~October 2026.

\bibitem{openaigym}
G.~Brockman \emph{et~al.}, ``OpenAI Gym,'' 2016, arXiv:1606.01540.

\bibitem{mujoco}
E.~Todorov, T.~Erez, and Y.~Tassa, ``MuJoCo: a physics engine for model-based control,'' in \emph{Proc. IEEE/RSJ Int. Conf. Intell. Robots Syst. (IROS)}, 2012, pp.~5026--5033, doi: \href{https://doi.org/10.1109/IROS.2012.6386109}{10.1109/IROS.2012.6386109}.

\bibitem{chen2020lbc}
D.~Chen, B.~Zhou, V.~Koltun, and P.~Kr\"{a}henb\"{u}hl, ``Learning by cheating,'' in \emph{Proc. Conf. Robot Learn. (CoRL)}, ser. Proc. Mach. Learn. Res., vol.~100, 2020, pp.~66--75. [Online]. Available: \url{https://proceedings.mlr.press/v100/chen20a.html}

\bibitem{cut}
T.~Park, A.A.~Efros, R.~Zhang, and J.-Y.~Zhu, ``Contrastive learning for unpaired image-to-image translation,'' in \emph{Computer Vision, ECCV 2020}, ser. Lect. Notes Comput. Sci., vol.~12354, 2020, pp.~319--345, doi: \href{https://doi.org/10.1007/978-3-030-58545-7_19}{10.1007/978-3-030-58545-7\_19}.

\bibitem{han2021dclgan}
J.~Han, M.~Shoeiby, L.~Petersson, and M.A.~Armin, ``Dual contrastive learning for unsupervised image-to-image translation,'' in \emph{Proc. IEEE/CVF Conf. Comput. Vis. Pattern Recognit. Workshops (CVPRW)}, 2021, pp.~746--755, doi: \href{https://doi.org/10.1109/CVPRW53098.2021.00084}{10.1109/CVPRW53098.2021.00084}.

\bibitem{kim2024unsb}
B.~Kim, G.~Kwon, K.~Kim, and J.C.~Ye, ``Unpaired image-to-image translation via neural Schr\"{o}dinger bridge,'' in \emph{Int. Conf. Learn. Represent. (ICLR)}, 2024, pp.~19312--19331. [Online]. Available: \url{https://arxiv.org/abs/2305.15086}

\bibitem{gonzalez2018}
R.C.~Gonzalez and R.E.~Woods, \emph{Digital Image Processing}, 4th~ed. Pearson, 2018.

\bibitem{stojanovski2023echo}
D.~Stojanovski, U.~Hermida, P.~Lamata, A.~Beqiri, and A.~Gomez, ``Echo from noise: synthetic ultrasound image generation using diffusion models for real image segmentation,'' in \emph{Simplifying Medical Ultrasound, ASMUS 2023}, ser. Lect. Notes Comput. Sci., vol.~14337, 2023, pp.~34--43, doi: \href{https://doi.org/10.1007/978-3-031-44521-7_4}{10.1007/978-3-031-44521-7\_4}.

\bibitem{zhou2022sdm}
W.~Zhou \emph{et~al.}, ``Semantic image synthesis via diffusion models,'' 2022, arXiv:2207.00050.

\bibitem{tomar2021conpres}
D.~Tomar, L.~Zhang, T.~Portenier, and O.~Goksel, ``Content-preserving unpaired translation from simulated to realistic ultrasound images,'' in \emph{Medical Image Computing and Computer Assisted Intervention, MICCAI 2021}, ser. Lect. Notes Comput. Sci., vol.~12908, 2021, pp.~659--669, doi: \href{https://doi.org/10.1007/978-3-030-87237-3_63}{10.1007/978-3-030-87237-3\_63}.

\bibitem{ronneberger2015unet}
O.~Ronneberger, P.~Fischer, and T.~Brox, ``U-Net: convolutional networks for biomedical image segmentation,'' in \emph{Medical Image Computing and Computer-Assisted Intervention, MICCAI 2015}, ser. Lect. Notes Comput. Sci., vol.~9351, 2015, pp.~234--241, doi: \href{https://doi.org/10.1007/978-3-319-24574-4_28}{10.1007/978-3-319-24574-4\_28}.

\bibitem{howard2019mobilenetv3}
A.~Howard \emph{et~al.}, ``Searching for MobileNetV3,'' in \emph{Proc. IEEE/CVF Int. Conf. Comput. Vis. (ICCV)}, 2019, pp.~1314--1324, doi: \href{https://doi.org/10.1109/ICCV.2019.00140}{10.1109/ICCV.2019.00140}.

\bibitem{chen2017deeplabv3}
L.-C.~Chen, G.~Papandreou, F.~Schroff, and H.~Adam, ``Rethinking atrous convolution for semantic image segmentation,'' 2017, arXiv:1706.05587.

\bibitem{he2016resnet}
K.~He, X.~Zhang, S.~Ren, and J.~Sun, ``Deep residual learning for image recognition,'' in \emph{Proc. IEEE Conf. Comput. Vis. Pattern Recognit. (CVPR)}, 2016, pp.~770--778, doi: \href{https://doi.org/10.1109/CVPR.2016.90}{10.1109/CVPR.2016.90}.

\bibitem{long2015fcn}
J.~Long, E.~Shelhamer, and T.~Darrell, ``Fully convolutional networks for semantic segmentation,'' in \emph{Proc. IEEE Conf. Comput. Vis. Pattern Recognit. (CVPR)}, 2015, pp.~3431--3440, doi: \href{https://doi.org/10.1109/CVPR.2015.7298965}{10.1109/CVPR.2015.7298965}.

\bibitem{isensee2021nnunet}
F.~Isensee, P.F.~Jaeger, S.A.A.~Kohl, J.~Petersen, and K.H.~Maier-Hein, ``nnU-Net: a self-configuring method for deep learning-based biomedical image segmentation,'' \emph{Nat. Methods}, vol.~18, pp.~203--211, 2021, doi: \href{https://doi.org/10.1038/s41592-020-01008-z}{10.1038/s41592-020-01008-z}.

\bibitem{kirillov2023sam}
A.~Kirillov \emph{et~al.}, ``Segment anything,'' in \emph{Proc. IEEE/CVF Int. Conf. Comput. Vis. (ICCV)}, 2023, pp.~3992--4003, doi: \href{https://doi.org/10.1109/ICCV51070.2023.00371}{10.1109/ICCV51070.2023.00371}.

\bibitem{ma2024medsam}
J.~Ma, Y.~He, F.~Li, L.~Han, C.~You, and B.~Wang, ``Segment anything in medical images,'' \emph{Nat. Commun.}, vol.~15, Art.~no.~654, 2024, doi: \href{https://doi.org/10.1038/s41467-024-44824-z}{10.1038/s41467-024-44824-z}.

\bibitem{lin2024samus}
X.~Lin, Y.~Xiang, L.~Yu, and Z.~Yan, ``Beyond adapting SAM: towards end-to-end ultrasound image segmentation via auto prompting,'' in \emph{Medical Image Computing and Computer Assisted Intervention, MICCAI 2024}, ser. Lect. Notes Comput. Sci., vol.~15008, 2024, pp.~24--34, doi: \href{https://doi.org/10.1007/978-3-031-72111-3_3}{10.1007/978-3-031-72111-3\_3}. Its preprint, arXiv:2309.06824, names the method SAMUS.

\bibitem{meyer2026ultrasam}
A.~Meyer, A.~Murali, F.~Zarin, D.~Mutter, and N.~Padoy, ``Ultrasam: a foundation model for ultrasound using large open-access segmentation datasets,'' \emph{Int. J. Comput. Assist. Radiol. Surg.}, vol.~21, pp.~93--102, 2026, doi: \href{https://doi.org/10.1007/s11548-025-03517-8}{10.1007/s11548-025-03517-8}.

\bibitem{jiao2024usfm}
J.~Jiao \emph{et~al.}, ``USFM: a universal ultrasound foundation model generalized to tasks and organs towards label efficient image analysis,'' \emph{Med. Image Anal.}, vol.~96, Art.~no.~103202, 2024, doi: \href{https://doi.org/10.1016/j.media.2024.103202}{10.1016/j.media.2024.103202}.

\bibitem{schulman2017ppo}
J.~Schulman, F.~Wolski, P.~Dhariwal, A.~Radford, and O.~Klimov, ``Proximal policy optimization algorithms,'' 2017, arXiv:1707.06347.

\bibitem{haarnoja2018sac}
T.~Haarnoja, A.~Zhou, P.~Abbeel, and S.~Levine, ``Soft actor-critic: off-policy maximum entropy deep reinforcement learning with a stochastic actor,'' in \emph{Proc. Int. Conf. Mach. Learn. (ICML)}, ser. Proc. Mach. Learn. Res., vol.~80, 2018, pp.~1861--1870. [Online]. Available: \url{https://proceedings.mlr.press/v80/haarnoja18b.html}

\bibitem{fujimoto2018td3}
S.~Fujimoto, H.~van~Hoof, and D.~Meger, ``Addressing function approximation error in actor-critic methods,'' in \emph{Proc. Int. Conf. Mach. Learn. (ICML)}, ser. Proc. Mach. Learn. Res., vol.~80, 2018, pp.~1587--1596. [Online]. Available: \url{https://proceedings.mlr.press/v80/fujimoto18a.html}

\bibitem{mnih2016a3c}
V.~Mnih \emph{et~al.}, ``Asynchronous methods for deep reinforcement learning,'' in \emph{Proc. Int. Conf. Mach. Learn. (ICML)}, ser. Proc. Mach. Learn. Res., vol.~48, 2016, pp.~1928--1937. [Online]. Available: \url{https://proceedings.mlr.press/v48/mniha16.html}

\bibitem{mnih2015dqn}
V.~Mnih \emph{et~al.}, ``Human-level control through deep reinforcement learning,'' \emph{Nature}, vol.~518, pp.~529--533, 2015, doi: \href{https://doi.org/10.1038/nature14236}{10.1038/nature14236}.

\bibitem{sb3}
A.~Raffin, A.~Hill, A.~Gleave, A.~Kanervisto, M.~Ernestus, and N.~Dormann, ``Stable-Baselines3: reliable reinforcement learning implementations,'' \emph{J. Mach. Learn. Res.}, vol.~22, no.~268, pp.~1--8, 2021. [Online]. Available: \url{https://jmlr.org/papers/v22/20-1364.html}

\bibitem{pomerleau1991}
D.A.~Pomerleau, ``Efficient training of artificial neural networks for autonomous navigation,'' \emph{Neural Comput.}, vol.~3, pp.~88--97, 1991, doi: \href{https://doi.org/10.1162/neco.1991.3.1.88}{10.1162/neco.1991.3.1.88}.

\bibitem{ross2011dagger}
S.~Ross, G.J.~Gordon, and J.A.~Bagnell, ``A reduction of imitation learning and structured prediction to no-regret online learning,'' in \emph{Proc. Int. Conf. Artif. Intell. Stat. (AISTATS)}, ser. Proc. Mach. Learn. Res., vol.~15, 2011, pp.~627--635. [Online]. Available: \url{https://proceedings.mlr.press/v15/ross11a.html}

\bibitem{rubinstein1999cem}
R.~Rubinstein, ``The cross-entropy method for combinatorial and continuous optimization,'' \emph{Methodol. Comput. Appl. Probab.}, vol.~1, pp.~127--190, 1999, doi: \href{https://doi.org/10.1023/A:1010091220143}{10.1023/A:1010091220143}.

\bibitem{wang2004ssim}
Z.~Wang, A.C.~Bovik, H.R.~Sheikh, and E.P.~Simoncelli, ``Image quality assessment: from error visibility to structural similarity,'' \emph{IEEE Trans. Image Process.}, vol.~13, pp.~600--612, 2004, doi: \href{https://doi.org/10.1109/TIP.2003.819861}{10.1109/TIP.2003.819861}.

\bibitem{studholme1999nmi}
C.~Studholme, D.L.G.~Hill, and D.J.~Hawkes, ``An overlap invariant entropy measure of 3D medical image alignment,'' \emph{Pattern Recognit.}, vol.~32, pp.~71--86, 1999, doi: \href{https://doi.org/10.1016/S0031-3203(98)00091-0}{10.1016/S0031-3203(98)00091-0}.

\bibitem{zhang2018lpips}
R.~Zhang, P.~Isola, A.A.~Efros, E.~Shechtman, and O.~Wang, ``The unreasonable effectiveness of deep features as a perceptual metric,'' in \emph{Proc. IEEE/CVF Conf. Comput. Vis. Pattern Recognit. (CVPR)}, 2018, pp.~586--595, doi: \href{https://doi.org/10.1109/CVPR.2018.00068}{10.1109/CVPR.2018.00068}.

\bibitem{kidneyus}
R.~Singla \emph{et~al.}, ``The open kidney ultrasound data set,'' in \emph{Simplifying Medical Ultrasound, ASMUS 2023}, ser. Lect. Notes Comput. Sci., vol.~14337, 2023, pp.~155--164, doi: \href{https://doi.org/10.1007/978-3-031-44521-7_15}{10.1007/978-3-031-44521-7\_15}.

\bibitem{aul}
Y.~Xu \emph{et~al.}, ``Improving artificial intelligence pipeline for liver malignancy diagnosis using ultrasound images and video frames,'' \emph{Brief. Bioinform.}, vol.~24, Art.~no.~bbac569, 2023, doi: \href{https://doi.org/10.1093/bib/bbac569}{10.1093/bib/bbac569}.

\bibitem{reinforce}
R.J.~Williams, ``Simple statistical gradient-following algorithms for connectionist reinforcement learning,'' \emph{Mach. Learn.}, vol.~8, pp.~229--256, 1992, doi: \href{https://doi.org/10.1007/BF00992696}{10.1007/BF00992696}.

\bibitem{kid}
M.~Bi\'{n}kowski, D.J.~Sutherland, M.~Arbel, and A.~Gretton, ``Demystifying MMD GANs,'' in \emph{Int. Conf. Learn. Represent. (ICLR)}, 2018. [Online]. Available: \url{https://arxiv.org/abs/1801.01401}

\bibitem{fid}
M.~Heusel, H.~Ramsauer, T.~Unterthiner, B.~Nessler, and S.~Hochreiter, ``GANs trained by a two time-scale update rule converge to a local Nash equilibrium,'' in \emph{Advances in Neural Information Processing Systems 30 (NIPS 2017)}, 2017. [Online]. Available: \url{https://arxiv.org/abs/1706.08500}

\bibitem{c2st}
D.~Lopez-Paz and M.~Oquab, ``Revisiting classifier two-sample tests,'' in \emph{Int. Conf. Learn. Represent. (ICLR)}, 2017. [Online]. Available: \url{https://arxiv.org/abs/1610.06545}

\bibitem{holm1979}
S.~Holm, ``A simple sequentially rejective multiple test procedure,'' \emph{Scand. J. Stat.}, vol.~6, pp.~65--70, 1979. [Online]. Available: \url{https://www.jstor.org/stable/4615733}

\bibitem{ji2022amos}
Y.~Ji \emph{et~al.}, ``AMOS: a large-scale abdominal multi-organ benchmark for versatile medical image segmentation,'' in \emph{Advances in Neural Information Processing Systems 35 (NeurIPS 2022), Datasets and Benchmarks Track}, 2022. [Online]. Available: \url{https://arxiv.org/abs/2206.08023}

\bibitem{shiralkar2015hri}
K.~Shiralkar, S.~Johnson, E.I.~Bluth, R.H.~Marshall, A.~Dornelles, and P.M.~Gulotta, ``Improved method for calculating hepatic steatosis using the hepatorenal index,'' \emph{J. Ultrasound Med.}, vol.~34, pp.~1051--1059, 2015, doi: \href{https://doi.org/10.7863/ultra.34.6.1051}{10.7863/ultra.34.6.1051}.

\bibitem{goksel2009}
O.~Goksel and S.E.~Salcudean, ``B-mode ultrasound image simulation in deformable 3-D medium,'' \emph{IEEE Trans. Med. Imaging}, vol.~28, pp.~1657--1669, 2009, doi: \href{https://doi.org/10.1109/TMI.2009.2016561}{10.1109/TMI.2009.2016561}.

\bibitem{burger2013}
B.~Burger, S.~Bettinghausen, M.~Radle, and J.~Hesser, ``Real-time GPU-based ultrasound simulation using deformable mesh models,'' \emph{IEEE Trans. Med. Imaging}, vol.~32, pp.~609--618, 2013, doi: \href{https://doi.org/10.1109/TMI.2012.2234474}{10.1109/TMI.2012.2234474}.

\bibitem{starkov2019}
R.~Starkov, L.~Zhang, M.~Bajka, C.~Tanner, and O.~Goksel, ``Ultrasound simulation with deformable and patient-specific scatterer maps,'' \emph{Int. J. Comput. Assist. Radiol. Surg.}, vol.~14, pp.~1589--1599, 2019, doi: \href{https://doi.org/10.1007/s11548-019-02054-5}{10.1007/s11548-019-02054-5}.

\bibitem{usbgymdata}
T.~Intharah, Z.~Gao, Y.~Liu, and H.~Dong, ``USB Gym: result files behind the tables and figures, version 1.0,'' Zenodo, 2026, dataset, license CC BY 4.0, doi: \href{https://doi.org/10.5281/zenodo.23913142}{10.5281/zenodo.23913142}.

\bibitem{auldata}
Y.~Xu \emph{et~al.}, ``Annotated Ultrasound Liver images,'' Zenodo, 2022, dataset, license CC BY 4.0, doi: \href{https://doi.org/10.5281/zenodo.7272660}{10.5281/zenodo.7272660}.

\bibitem{vitaledata}
S.~Vitale and J.I.~Orlando, ``US simulation \& segmentation: real and synthetic abdominal ultrasound scans, with manual segmentations,'' Kaggle, 2020, dataset. [Online]. Available: \url{https://www.kaggle.com/datasets/ignaciorlando/ussimandsegm}

\bibitem{claudecode}
Anthropic, ``Claude Code,'' AI coding assistant, models Claude Opus 4.8, Claude Opus 5 and Claude Opus 5.5, 2026. [Online]. Available: \url{https://claude.com/claude-code}

\ifdefined\usbgsupplement

\bibitem{seabold2010statsmodels}
S.~Seabold and J.~Perktold, ``Statsmodels: econometric and statistical modeling with Python,'' in \emph{Proc. 9th Python in Science Conf. (SciPy 2010)}, 2010, pp.~92--96, doi: \href{https://doi.org/10.25080/Majora-92bf1922-011}{10.25080/Majora-92bf1922-011}.

\bibitem{viola1997mi}
P.~Viola and W.M.~Wells~III, ``Alignment by maximization of mutual information,'' \emph{Int. J. Comput. Vis.}, vol.~24, pp.~137--154, 1997, doi: \href{https://doi.org/10.1023/A:1007958904918}{10.1023/A:1007958904918}.

\bibitem{maes1997mi}
F.~Maes, A.~Collignon, D.~Vandermeulen, G.~Marchal, and P.~Suetens, ``Multimodality image registration by maximization of mutual information,'' \emph{IEEE Trans. Med. Imaging}, vol.~16, pp.~187--198, 1997, doi: \href{https://doi.org/10.1109/42.563664}{10.1109/42.563664}.

\bibitem{adam}
D.P.~Kingma and J.~Ba, ``Adam: a method for stochastic optimization,'' in \emph{Int. Conf. Learn. Represent. (ICLR)}, 2015. [Online]. Available: \url{https://arxiv.org/abs/1412.6980}

\fi
\end{thebibliography}
\end{document}